%% file: 0main.tex
\documentclass{article} 
\usepackage{iclr,times}
\usepackage{makecell}
\usepackage{amsmath,amssymb} 
\usepackage[ruled,vlined]{algorithm2e}
\usepackage{color}
\usepackage{xcolor,colortbl}
\definecolor{Gray}{gray}{0.9}
\usepackage{graphicx}
\usepackage{wrapfig}
\usepackage{xspace}
\usepackage{enumitem,booktabs}
\usepackage{footnote}
\usepackage{graphicx}
\graphicspath{
{figs/},
{figs/main_table/},
{figs/teaser/}
}

\input{math_commands.tex}

\usepackage{hyperref}
\usepackage{url}

\newcommand{\eg} {{\it e.g.,} }

\def\fig#1{Fig.~\ref{fig:#1}}

\newcommand{\algorithmShort}{SPML\xspace}
\newcommand{\lambdaImgSim}{\lambda_{I}}
\newcommand{\lambdaSemAnn}{\lambda_{C}}
\newcommand{\lambdaSemCoc}{\lambda_{O}}
\newcommand{\lambdaFeatAff}{\lambda_{A}}
\newcommand{\concImgSim}{\kappa_{I}}
\newcommand{\concSemAnn}{\kappa_{C}}
\newcommand{\concSemCoc}{\kappa_{O}}
\newcommand{\concFeatAff}{\kappa_{A}}

\let\oldnl\nl
\newcommand{\nonl}{\renewcommand{\nl}{\let\nl\oldnl}}

\newcommand{\tb}[3]{\setlength{\tabcolsep}{#2mm}\begin{tabular}{@{}#1@{}} #3\end{tabular}}

\def\imwh#1#2#3{\includegraphics[width=#2\textwidth,height=#3\textheight]{#1}}

\title{
Universal Weakly Supervised Segmentation \\
by Pixel-to-Segment Contrastive Learning
}

\author{%
Tsung-Wei Ke \hspace{15pt}
Jyh-Jing Hwang \hspace{15pt}
Stella X. Yu \\
UC Berkeley / ICSI\\
\texttt{\{twke,jyh,stellayu\}@berkeley.edu}
}

\iclrfinalcopy 
\begin{document}

\maketitle

\input{1abs.tex}
\input{2intro.tex}
\input{3work.tex}
\input{4method.tex}
\input{5exp.tex}

\input{6conclusion.tex}

{\bf Acknowledgements.} 
This work was supported, in part, by Berkeley Deep Drive and Berkeley AI Research Commons with Facebook. This work used the Extreme Science and Engineering Discovery Environment (XSEDE), which is supported by National Science Foundation grant number ACI-1548562. Specifically, it used the Bridges system, which is supported by NSF award number ACI-1445606, at the Pittsburgh Supercomputing Center (PSC).

\newpage
\bibliography{iclr}
\bibliographystyle{iclr}

\newpage
\input{7appendix.tex}

\end{document}

%% file: math_commands.tex
\usepackage{amsmath,amsfonts,bm}

\def\eqref#1{equation~\ref{#1}}

\def\1{\bm{1}}

\DeclareMathAlphabet{\mathsfit}{\encodingdefault}{\sfdefault}{m}{sl}
\SetMathAlphabet{\mathsfit}{bold}{\encodingdefault}{\sfdefault}{bx}{n}

\DeclareMathOperator*{\argmax}{arg\,max}

%% file: 1abs.tex
\begin{abstract}

Weakly supervised segmentation requires assigning a label to every pixel based on training instances with partial annotations such as image-level tags, object bounding boxes, labeled points and scribbles.  
This task is challenging, as coarse annotations ({\it tags}, {\it boxes}) lack precise pixel localization whereas sparse annotations ({\it points}, {\it scribbles}) lack broad region coverage.
Existing methods tackle these two types of weak supervision differently: Class activation maps are used to localize coarse labels and iteratively refine the segmentation model, whereas conditional random fields are used to propagate sparse labels to the entire image. 

We formulate weakly supervised segmentation as a semi-supervised metric learning problem, where pixels of the same (different) semantics need to be mapped to the same (distinctive) features.  
We propose 4 types of contrastive relationships between pixels and segments in the feature space, capturing low-level image similarity, semantic annotation, co-occurrence, and feature affinity.  They act as priors; the pixel-wise feature can be learned from training images with any partial annotations in a data-driven fashion.
In particular, unlabeled pixels in training images participate not only in data-driven grouping within each image, but also in discriminative feature learning {\it within} and {\it across} images. 
We deliver a universal weakly supervised segmenter with significant gains on Pascal VOC and DensePose. Our code is publicly available at \url{https://github.com/twke18/SPML}.
\end{abstract}

%% file: 2intro.tex
\def\imrow#1{
\imwh{Point#1.png}{0.195}{0.07}&
\imwh{Image#1.jpg}{0.195}{0.07}&
\imwh{Base#1.png}{0.195}{0.07}&
\imwh{Spml#1.png}{0.195}{0.07}&
\imwh{Gt#1.png}{0.195}{0.07}\\[-2pt]
}
\def\imrowT#1{
&
\imwh{Image#1.jpg}{0.195}{0.07}&
\imwh{Base#1.png}{0.195}{0.07}&
\imwh{Spml#1.png}{0.195}{0.07}&
\imwh{Gt#1.png}{0.195}{0.07}\\[-2pt]
}
\def\figTeaserDensePose#1{
\begin{figure}[#1]\centering
\tb{ccccc}{0.2}{
\imrow{1}
\imrow{2}
\bf weak supervision &
\bf image &
\bf baseline&
\bf ours&
\bf ground-truth\\
\imrowT{6}
}
\caption{
Our task learns a segmenter given partially labeled training images and applies it to test images.  A common baseline is to propagate labels within an image based on feature similarity.  We model it as semi-supervised metric learning and learn the pixel-wise feature by contrasting it within and across images.  Our results are fuller and more accurate, approaching the ground-truth.}
    \label{fig:teaser_densepose}
\end{figure}
}

\def\imrrow#1{
\imwh{tableImage#1.png}{0.19}{0.07}&
\imwh{tableTag#1.png}{0.19}{0.07}&
\imwh{tableBox#1.png}{0.19}{0.07}&
\imwh{tablePoint#1.png}{0.19}{0.07}&
\imwh{tableScribble#1.png}{0.19}{0.07}\\
}
\def\figTeaser#1{
\begin{figure}[#1]\centering
    \tb{c|c|c|c|c}{0.5}{
    \toprule
    \bf{image} & \bf{image tags} & \bf{bounding boxes} & \bf{labeled points} & \bf{scribbles}\\
    \midrule
    \imrrow{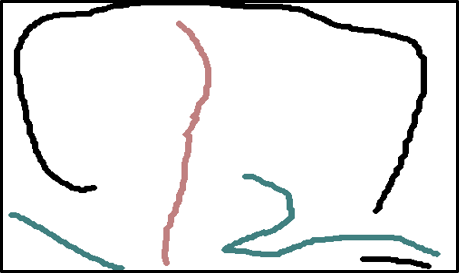}
    \midrule
    \bf SOTA methods & CAM + refine & box-wise CAM & CRF loss &  CRF loss\\
    \midrule
    \bf our method & \multicolumn{4}{c}{single pixel-to-segment contrastive learning loss formulation}\\
    \midrule
    \bf our relative gain & $+8.6\%$   &  $+4.7\%$ &   $+24.7\%$ &  $+1.4\%$\\
    \bottomrule
    }
    \caption{We propose a unified framework for weakly supervised semantic segmentation with different types of annotations.  We demonstrate consistent performance gains compared to the state-of-the-art (SOTA) methods: ~\cite{chang2020weak} for image tags, ~\cite{song2019box} for bounding boxes, and ~\cite{tang2018regularized} for points and scribbles. For tags and boxes, Class Activation Maps (CAM) ~\citep{zhou2016learning} are often used to localize semantics as an initial mask and iteratively refine the segmentation model, whereas for labeled points and scribbles, Conditional Random Fields (CRF) are used to propagate semantic labels to unlabeled regions based on low-level image similarity.}
    \label{fig:teaser}
\end{figure}
}

\def\figTeaserDensePoseOld#1{
\begin{figure}[#1]
    \centering
    \includegraphics[width=\textwidth]{figs/teaser_denspose.png}
    \caption{
    Sample weakly supervised segmentation task and result comparisons.  Given an image (a) and a single seed pixel labeling per body part (b), the goal is to segment out all the body parts (e).  The common baseline approach (c) is to propagate seed pixel labels within each image based on  feature similarity.  Our approach (d) treats segmentation as a semi-supervised pixel-wise metric learning problem, propagates labels by data-driven grouping, learns the feature by contrasting it within and across images.  Our segmentation is fuller and more accurate, approaching the ground-truth.}
    \label{fig:teaser_densepose}
\end{figure}
}

\def\figTeaserOld#1{
\begin{figure}[#1]
    \includegraphics[clip,width=0.99\textwidth]{figs/teaser.png}
    \caption{We propose a unified framework for weakly supervised semantic segmentation with different types of annotations.  We demonstrate consistent performance gains compared to the state-of-the-art (SOTA) methods: ~\cite{chang2020weak} for image tags, ~\cite{song2019box} for bounding boxes, and ~\cite{tang2018regularized} for points and scribbles. For tags and boxes, Class Activation Maps (CAM) ~\citep{zhou2016learning} are often used to localize semantics as an initial mask and iteratively refine the segmentation model, whereas for labeled points and scribbles, Conditional Random Fields (CRF) are used to propagate semantic labels to unlabeled regions based on low-level image similarity.}
  \vspace{-6pt}
    \label{fig:teaser}
\end{figure}
}

\figTeaserDensePose{!b}

\section{Introduction}
\label{sec:intro}

Consider the task of learning a semantic segmenter given sparsely labeled training images (Fig.~\ref{fig:teaser_densepose}): Each body part is labeled with a single seed pixel and the task is to segment out the entire person by individual body parts, even though the ground-truth segmentation is not known during training.
This task is challenging, as not only a single body part could contain several visually distinctive areas (\eg {\it head} consists of {\it eyes}, {\it nose}, {\it mouth}, {\it beard}), but two adjacent body parts could also have the same visual appearance (\eg  {\it upper arm}, {\it lower arm}, and {\it hand} have the same skin appearance).  Once the segmenter is learned, it can be applied to a test image without any annotations.

\figTeaser{!t}

This task belongs to a family of weakly supervised segmentation problems, the goal of which is to assign a label to each pixel despite that only partial supervision is available during training.  It addresses the practical issue of learning segmentation from  minimum annotations.
Such weak supervision takes many forms, e.g., image tags \citep{kolesnikov2016seed,ahn2018learning,huang2018weakly,lee2019ficklenet}, bounding boxes~\citep{dai2015boxsup,khoreva2017simple,song2019box},  keypoints~\citep{bearman2016s}, and scribbles~\citep{lin2016scribblesup,tang2018normalized,tang2018regularized}.  Tags and boxes are coarse annotations that lack precise pixel localization whereas  points and scribbles are sparse annotations that lack broad region coverage.

Weakly supervised semantic segmentation can be regarded as a semi-supervised pixel classification problem: Some pixels or pixel sets have labels, most don't, and the key is how to propagate and refine annotations from coarsely and sparsely labeled pixels to unlabeled pixels.  

Existing methods tackle two types of weak supervision differently: Class Activation Maps (CAM) ~ \citep{zhou2016learning} are used to localize coarse labels, generate pseudo pixel-wise labels, and iteratively refine the segmentation model, whereas Conditional Random Fields (CRF) ~\citep{krahenbuhl2011efficient} are used to propagate sparse labels to the entire image.
These ideas can be incorporated as an additional unsupervised loss on the feature learned for segmentation~\citep{tang2018regularized}: While labeled pixels receive supervision, unlabeled pixels in different segments shall have distinctive feature representations.  

We propose a {\it
\underline{S}emi-supervised \underline{P}ixel-wise  \underline{M}etric \underline{L}earning} (SPML) model that can  handle all these weak supervision varieties with {a single pixel-to-segment contrastive learning formulation} (Fig.~\ref{fig:teaser}).
Instead of classifying pixels, our metric learning model learns a pixel-wise feature embedding based on common grouping relationships that can be derived from any form of weak supervision.

Our key insight is to integrate unlabeled pixels into both supervised labeling and discriminative feature learning.  They shall participate not only in data-driven grouping within each image, but also in discriminative feature learning {\it within} and more importantly {\it across} images. 
Intuitively, labeled pixels receive supervision not only for themselves,  but also for their surround pixels that share visual similarity. 
On the other hand, unlabeled pixels are not just passively brought into discriminative learning induced by sparsely labeled pixels, they themselves are organized based on bottom-up grouping cues (such as grouping by color similarity and separation by strong contours).  When they are examined {\it across} images, repeated patterns of frequent occurrences would also form a cluster that demand active discrimination from other patterns.

We capture the above insight in a single pixel-wise metric learning objective for segmentation, the goal of which is to map each pixel into a point in the feature space so that pixels in the same (different) semantic groups are close (far) in the feature space.
Our model extends SegSort~\citep{hwang2019segsort} from its fully supervised and unsupervised segmentation settings to a universal weakly-supervised segmentation setting.   With a single consistent feature learning criterion, such a model sorts pixels discriminatively within individual images and sorts segment clusters discriminatively across images, both steps minimizing the same feature discrimination loss.

Our experiments on 
Pascal VOC~\citep{everingham2010pascal} and  DensePose~\citep{alp2018densepose}
demonstrate consistent gains over the state-of-the-art (SOTA), and the gain is substantial especially for the sparsest keypoint supervision.

%% file: 3work.tex
\section{Related Work}
\label{sec:work}

\noindent {\bf Semi-supervised learning.}
~\cite{weston2012deep} treats it as a joint learning problem with both labeled and unlabeled data.
One way is to capture the underlying structure of unlabeled data with generative models~\citep{kingma2014semi,rasmus2015semi}.
%
Another way is to regularize feature learning through a consistency loss,  \eg adversarial ensembling~\citep{miyato2018virtual}, imitation learning and distillation~\citep{tarvainen2017mean}, cross-view ensembling~\citep{clark2018cross}.
These methods are most related to transductive learning~\citep{joachims2003transductive,zhou2004learning,fergus2009semi,liu2019deep}, where labels are propagated to unlabeled data via clustering in the pre-trained feature space.
Our work does transductive learning in an adaptively learned feature space.

\noindent {\bf Weakly-supervised semantic segmentation.} 
Partial annotations include scribbles~\citep{lin2016scribblesup,tang2018normalized,tang2018regularized,wang2019boundary}, bounding boxes~\citep{dai2015boxsup,khoreva2017simple,song2019box}, points~\citep{bearman2016s}, or image tags~\citep{papandreou2015weakly,kolesnikov2016seed,ahn2018learning,huang2018weakly,li2018tell,lee2019ficklenet,shimoda2019self,zhang2019reliability,yao2020saliency,chang2020weak,araslanov2020single,wang2020self,Fan_2020_CVPR,sun2020mining}.  ~\cite{xu2015learning} formulates all  types of weak supervision as linear constraints on a SVM. ~\cite{papandreou2015weakly} bootstraps segmentation predictions via EM-optimization. Recent works \citep{lin2016scribblesup,kolesnikov2016seed,pathak2015constrained} typically use CAM~\citep{zhou2016learning} to obtain an initial dense mask and then train a model iteratively. GAIN~\citep{li2018tell} utilizes image tags or bounding boxes to refine these class-specific activation maps. ~\cite{sun2020mining} considers within-image relationships and explores the idea of co-segmentation. ~\cite{Fan_2020_CVPR} estimates the foreground and background for each category, with which the network learns to generate more precise CAMs. Regularization is enforced  at either the image level~\citep{lin2016scribblesup,kolesnikov2016seed,pathak2015constrained} or the feature level~\citep{tang2018normalized,tang2018regularized} to produce better dense masks. 
We incorporate this concept into adaptive feature learning and train the model only once.  All types of weak annotations are dealt with in a single contrastive learning framework.

\noindent {\bf Non-parametric  segmentation.}  Prior to deep learning, non-parametric models~\citep{russell2009segmenting,tighe2010superparsing,liu2011nonparametric} usually use designed features with statistical or graphical models to segment images.
Recently, inspired by non-parametric models for  recognition~\citep{wu2018unsupervised,wu2018improving}, SegSort~\citep{hwang2019segsort} captures pixel-to-segment relationships via a pixel-wise embedding and develops the first deep non-parametric semantic segmentation for supervised and unsupervised settings.  
Building upon SegSort, our work has the flexibility of a non-parametric model at capturing data relationships  and modeling subclusters within a category.

%% file: 4method.tex
\def\figMain#1{
\begin{figure}[#1]
    \centering
    \vspace{-24pt}
    \includegraphics[width=0.95\textwidth]{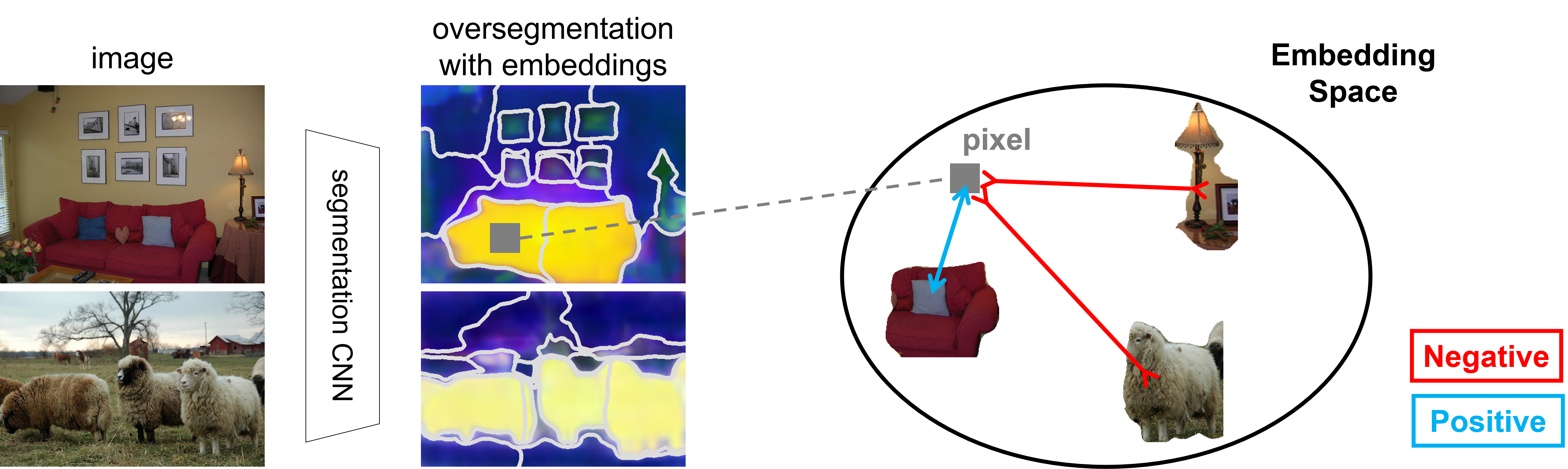}
    \caption{Overall method diagram.  We develop pixel-wise embeddings with contrastive learning between pixels and segments. We derive various forms of positive and negative segments for each pixel. Our goal is to \textcolor{blue}{ attract (blue inward arrows)} the pixel with positive segments, while \textcolor{red}{ repelling (red outward arrows)} it from negative segments in the feature space.  }
    \label{fig:Main}
\end{figure}
}

\def\figRelationships#1{
\begin{figure}[#1]\centering
\includegraphics[width=\textwidth]{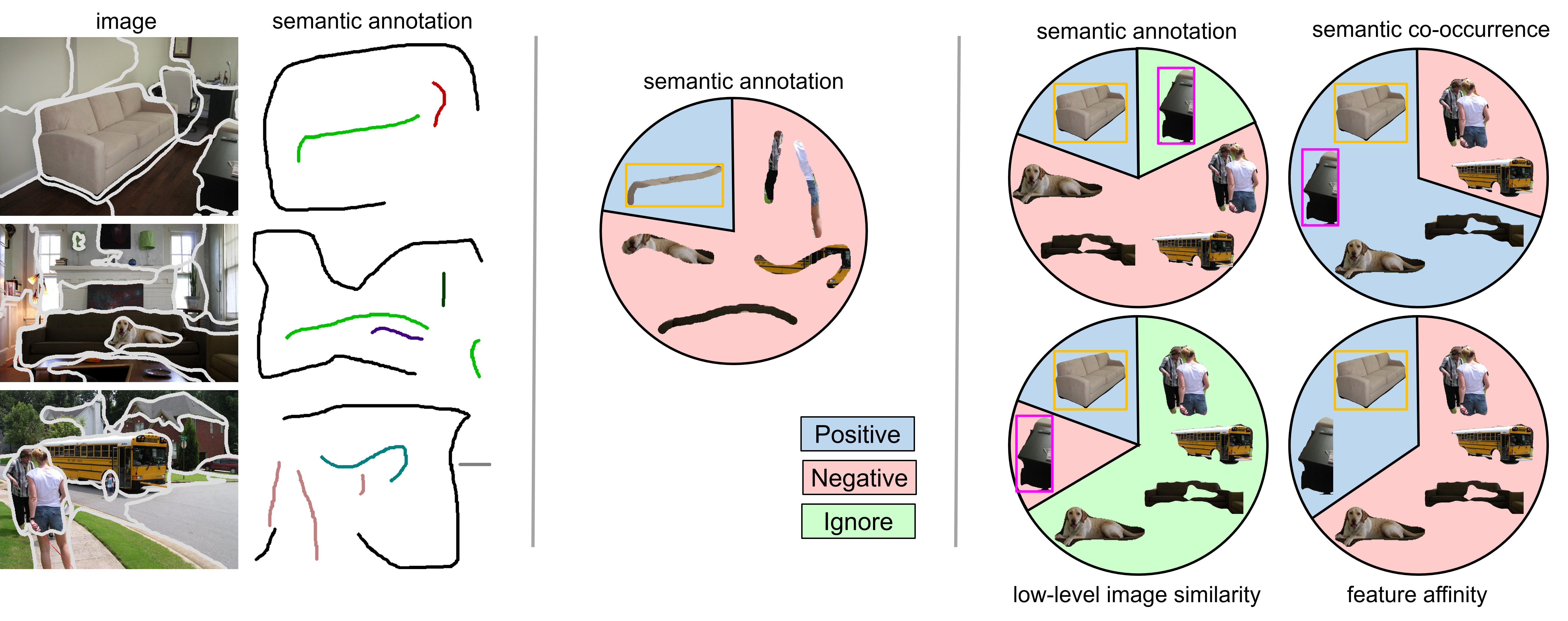}\\
\tb{ccc}{10}{
{\bf a)} training data &
{\bf b)} existing methods &
{\bf c)} our SPML\\
}    
\caption{
Our method uses labeled and unlabeled portions of the training data more extensively.
{\bf a)} Training images and their labeled scribbles are sparse and incomplete. {\bf b)} Existing methods train a pixel-wise classifier using only labeled pixels and propagate labels within each image. 
{\bf c)} Our method leverages {\it four} types of pixel-to-segment semantic relationships to augment the labeled sets, includes unlabeled pixels (fuller segments than just thin scribbles) and unlabeled segments (e.g. {\it \textcolor{magenta}{desk}} outlined in magenta), forms dynamic contrastive relationships between segments (e.g. the {\it \textcolor{magenta}{desk}} can be positive, negative, or to be ignored to the  {\it \textcolor{orange}{sofa}} in different relations.
}
    \label{fig:Relationships}
\end{figure}
}

\def\figVisRelations#1{
\begin{figure}[#1]
    \centering
    \includegraphics[width=\textwidth]{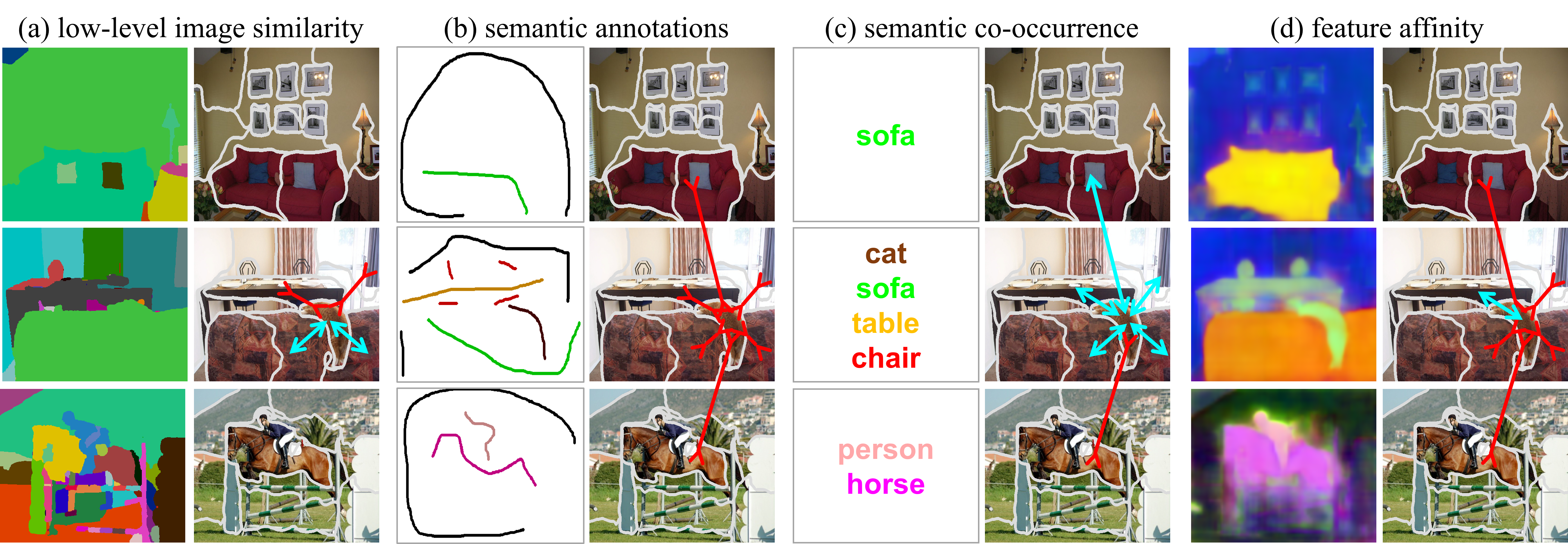}
    \vspace{-10pt}
    \caption{Four types of pixel-to-segment attraction and repulsion relationships.
     A pixel is \textcolor{blue}{attracted to} (\textcolor{red}{repelled by}) segments:
    {\bf a)} of similar (different) visual appearances such as color or texture,
    {\bf b)} of the same (different) class labels,
    {\bf c)} in images with common (distinctive) labels,
    {\bf d)} of nearby (far-away) feature embeddings.
    They form different positive and negative sets.
    }
    \label{fig:VisRelations}
\end{figure}
}

\figMain{!t}
\figVisRelations{!t}

\section{Semi-Supervised Pixel-wise Metric Learning Method}
\label{sec:method}

Metric learning develops a feature representation based on data grouping and separation cues.  Our method (Fig.~\ref{fig:Main}) segments an image by learning a pixel-wise embedding with a contrastive loss between pixels and segments:
For each pixel $i$, we learn a
latent feature $\phi(i)$ such that $i$ is close to its positive segments (exemplars) and far from its negative ones in that feature space.  

In the fully supervised setting, we can define pixel $i$'s positive and negative sets, denoted by 
$\mathcal{C}^+$ and $\mathcal{C}^-$ respectively,
as pixels in the same (different) category.  However, this idea is not applicable to weakly- or un-supervised settings where the label is not available on every pixel.  In the labeled points setting, $\mathcal{C}^+$ and $\mathcal{C}^-$ would only contain a few exemplars according to the sparse pixel labels.

Our basic idea is to enlarge the sets of $\mathcal{C}^+$ and $\mathcal{C}^-$ to improve the feature learning efficacy.  By exploring different relationships and assumptions in the image data, we are able to generate abundant positive and negative segments for any pixel at the same time, providing more supervision in the latent feature space.  We propose four types of  relationships between pixels and segments (Fig.~\ref{fig:VisRelations}):
\begin{enumerate}[leftmargin=*,itemsep=0pt]
    \item \textbf{Low-level image similarity}: 
    We impose a spatial smoothness prior on the pixel-wise feature to keep pixels together in visually coherent regions.  The segment pixel $i$ belongs to based on low-level image cues is a positive segment to pixel $i$; any other segments are negative ones.
    \item \textbf{Semantic annotation}: 
    We expand the semantics from labeled points and  scribbles to  pseudo-labels inferred from image- or box-wise CAM.  The label of a segment can be estimated by majority vote among pixels; if it is the same as pixel $i$'s, the segment is a positive segment to $i$.
    
    \item \textbf{Semantic co-occurrence}:
    We expand the semantics by assuming that pixels in similar semantic contexts tend to be grouped together. If a segment appears in an image that shares any of the semantic classes as pixel $i$'s image, it is a positive segment to $i$ and otherwise a negative one.
    \item \textbf{Feature affinity}: 
    We impose a featural smoothness prior assuming that pixels and segments of the same semantics form a cluster in the feature space.  We propagate the semantics within and across images from pixel $i$ to its closest segment $s$ in the feature space.
\end{enumerate}



\subsection{Pixel-to-Segment Contrastive Grouping Relationships}
\label{subsec:method_relationships}

Our goal is to propagate known semantics from labeled data $\mathcal{C}$ to unlabeled data $\mathcal{U}$ with the aforementioned priors.  
$\mathcal{C}$ and $\mathcal{U}$ denote the sets of segment indices respectively.  We detail how to augment positive / negative segment sets using both $\mathcal{C}$ and $\mathcal{U}$ for each type of relationships (\fig{VisRelations}).

\noindent {\bf Low-level image similarity.}
To propagate labels within visually coherent regions, we generate a low-level over-segmentation.  Following SegSort~\citep{hwang2019segsort}, we use the HED contour detector~\citep{xie2015holistically} (pre-trained on BSDS500 dataset~\citep{arbelaez2010contour}) and gPb-owt-ucm~\citep{arbelaez2010contour} to generate a  segmentation without semantic information. 
%
We define $i$'s positive and negative segments as $i$'s own segment and all the other segments, denoted as $\mathcal{V^+}$ and $\mathcal{V^-}$ respectively.
%
%
%
We only consider segments in the same image as pixel $i$'s.  We align the contour-based over-segmentations with segmentations generated by K-Means clustering as in SegSort.

\noindent {\bf Semantic annotation.}
Image tags and bounding boxes do not provide pixel-wise localization.  We derive pseudo labels from image- or box-wise CAM and align them with oversegmentations induced by the pixel-wise feature.  Pixel $i$'s positive (negative) segments are the ones with the same (different) semantic category, denoted by 
$\mathcal{C^+}$ and $\mathcal{C^-}$ respectively.
We ignore all the unlabeled segments. 
%
%

\noindent {\bf Semantic co-occurrence.} 
Semantic context characterizes the co-occurrences of different objects, which can be used as a prior to group and separate pixels.  We define semantic context as the union of object classes in each image.  Even without the pixel-wise localization of semantic labels, we can leverage semantic context to impose global regularization on the latent feature: The feature should separate images without any overlapping object categories.

Let $\mathcal{O^+}$ ($\mathcal{O^-}$) denote the set of segments in images with (without) overlapping categories as pixel $i$'s image.  That is,  if the image of pixel $i$ and another image share any semantic labels (\fig{VisRelations}c: {\it \{cat, sofa, table, chair\} } for the pixel in the Row 2 image vs. {\it \{sofa\}} for the Row 1 image), then all the segments from that image are positive segments to $i$ and included in $\mathcal{O^+}$; otherwise they are considered negative segments in $\mathcal{O^-}$  (\fig{VisRelations}c: all the segments in the Row 3 image).  In particular, all the segments in pixel $i$'s image are in $\mathcal{O^+}$ of $i$.
%
This semantic context relationship does not require localized annotations yet imposes regularization on pixel feature learning.

\noindent {\bf Feature affinity.}
Our goal is to learn a pixel-wise feature that indicates semantic segmentation.  It is thus reasonable to assume that pixels and segments of the same semantics form a cluster in the feature space, and we reinforce such clusters with a featural smoothness prior:
We find nearest neighbours in the feature space and propagate labels accordingly.


%
%
Specifically, we assign a semantic label to each unlabeled segment by finding its nearest labeled segment in the feature space.  We denote this expanded labeled set by $\hat{\mathcal{C}}$.  For pixel $i$, we define its positive (negative) segment set $\hat{\mathcal{C}}^+$ ($\hat{\mathcal{C}}^-$) according to whether a segment has the same label as $i$.

Our feature affinity relationship works best when: 1) the original labeled set is large enough to cover the feature space, 2) the labeled segments are distributed uniformly in the feature space, and 3) the pixel-wise feature already encodes certain semantic information.  We thus only apply to DensePose keypoint annotations in our experiments, where each body part is annotated by a point.

\figRelationships{!t}

\subsection{Pixel-wise Metric Learning Loss}
\label{subsec:method_segsort}
SegSort~\citep{hwang2019segsort} is an end-to-end segmentation model that generates a pixel-wise feature map and a resulting segmentation.  
Assuming independent normal distributions for individual segments, SegSort seeks a maximum likelihood estimation of the feature mapping, so that the feature induced partitioning in the image and clustering across images provide maximum discrimination among segments.
During inference, the segment label is predicted by K-Nearest Neighbor retrievals.

The feature induced partitioning in each image is calculated via spherical K-Means clustering~\citep{banerjee2005clustering}.
Let $\pmb{e}_i$ denote the feature vector at pixel $i$, which contains the mapped feature $\phi(i)$ and $i$'s spatial coordinates.  Let $z_i$ denote the index of the segment that pixel $i$ belongs to, $\pmb{R}_s$ the set of pixels in segment $s$, and $\pmb{\mu}_s$ the segment feature calculated as the spherical cluster centroid of segment $s$.
In the  Expectation-Maximization (EM) procedure for spherical K-means, the E-step calculates the most likely segment pixel $i$ belongs to: 
$z_i = \argmax_s \pmb{\mu}_s' {\pmb{e}_i}$,
and the M-Step updates the segment feature as the mean pixel-wise feature:
$\pmb{\mu}_s = \frac{\sum_{i \in \pmb{R}_s} \pmb{e}_i}{\|\sum_{i \in \pmb{R}_s} \pmb{e}_i\|}$.

Let $s$ denote the resulting segment that pixel $i$ belongs to per spherical clustering.  The posterior probability of pixel $i$ in segment $s$ can be evaluated over the set of all segments $S$ as:
\begin{align}
p(z_i=s|\pmb{e}_i, \pmb{\mu}) = \frac{\exp(\kappa\, \pmb{\mu}_s'\, \pmb{e}_i)}{\sum_{t \in S} \exp(\kappa \,\pmb{\mu}_t'\, \pmb{e}_i)}
\end{align}
where $\kappa$ is a concentration hyper-parameter.  SegSort minimizes the negative log-likelihood loss:
\begin{align}
    L_{\mathrm{SegSort}}(i) = -\log p(z_i=s|\pmb{e}_i, \pmb{\mu})
    = -\log \frac{\exp(\kappa\, \pmb{\mu}_s' \,\pmb{e}_i)}{\sum_{t \in S} \exp(\kappa\, \pmb{\mu}_t'\, \pmb{e}_i)}.
    \label{eqn:loss_segsort}
\end{align}
SegSort adopts soft neighborhood assignment \citep{goldberger2005neighbourhood}
to further strengthen the grouping of same-category segments.
Let $\mathcal{C}^+$ ($\mathcal{C}^-$) denote the index set of segments in the same (different) category as pixel $i$ except $s$ -- the segment $i$ belongs to.  We have:
\begin{align}
    L_{\mathrm{SegSort^+}}(i, \mathcal{C}^+, \mathcal{C}^-) = -\log \sum_{t \in \mathcal{C}^+} p(z_i=t|\pmb{e}_i, \pmb{\mu})
    = -\log \frac{\sum_{t \in \mathcal{C}^+} \exp(\kappa\, \pmb{\mu}_t'\, \pmb{e}_i)}{\sum_{t \in \mathcal{C}^+ \cup \mathcal{C}^-} \exp(\kappa\, \pmb{\mu}_t' \, \pmb{e}_i)}.
    \label{eqn:loss_segsort_neighbor}
\end{align}
For our weakly supervised segmentation, the total pixel-to-segment contrastive loss for pixel $i$ consists of 4 terms, one for each of the 4 pixel-to-segment attraction and repulsion relationships:
\begin{align}
    L(i) &= \lambdaImgSim L_{\mathrm{SegSort^+}}(i, \mathcal{V}^+, \mathcal{V}^-) + \lambdaSemAnn L_{\mathrm{SegSort^+}}(i, \mathcal{C}^+, \mathcal{C}^-)\nonumber\\
    &+ \lambdaSemCoc L_{\mathrm{SegSort^+}}(i, \mathcal{O}^+, \mathcal{O}^-) + \lambdaFeatAff L_{\mathrm{SegSort^+}}(i, \hat{\mathcal{C}}^+, \hat{\mathcal{C}}^-),
    \label{eqn:loss_augmented}
\end{align}
where $\lambdaSemAnn=1$.  Fig.~\ref{fig:Relationships} shows how our metric learning method 
utilizes labeled and unlabeled pixels and segments more extensively than existing classification methods: Our pseudo-labeled sets are fuller than labeled thin scribbles and include unlabeled segments; there are 3 more relationships other than semantic annotations; our segments participate in contrastive learning with dynamic roles in different relations.  By easily integrating a full range of pixel-to-segment attraction and repulsion relationships from low-level image similarity to mid-level feature affinity, and to high-level semantic co-occurrence, we go far beyond the direct supervision from semantic annotations.

%% file: 5exp.tex
\section{Experiments on PASCAL VOC and DensePose}
\label{sec:exp}

\noindent \textbf{Datasets.}
{\bf Pascal VOC 2012} ~\cite{everingham2010pascal} includes 20 object categories and one background class. Following ~\cite{chen2017deeplab}, we use the augmented training set with 10,582 images and validation set with 1,449 images. We use the scribble annotations provided by ~\cite{lin2016scribblesup} for training.
{\bf DensePose}~\citep{alp2018densepose} is a human pose parsing dataset based on MSCOCO~\citep{lin2014microsoft}. The dataset is annotated with 14 body part classes. We extract the keypoints from the center of each part segmentation. The training set includes 26,437 images and we use minival2014 set for testing, which includes 1,508 images. See Appendix for more details.

\begin{table}
\begin{center}
\setlength{\tabcolsep}{10pt}
\begin{tabular}{c|c|c|c|c}
\toprule
    Dataset & Annotation & $\lambdaImgSim$ & $\lambdaSemCoc$ & $\lambdaFeatAff$\\
    \midrule
    Pascal & image tags & 0.3 & 1.0 & 0.0\\
    & boxes & 0.3 & 1.0 & 0.0\\
    & points & 1.0 & 1.0 & 0.0\\
    & scribbles & 0.1 & 0.5 & 0.0\\
    \midrule
    DensePose & points & 0.1 & 0.0 & 0.5\\
    \bottomrule
\end{tabular}
\end{center}
\vspace{-12pt}
\caption{Hyper-parameters for different types of annotations on Pascal and DensePose dataset.}
\label{tab:annotations}
\end{table}

\noindent \textbf{Architecture, training and testing.}
For all the experiments on PASCAL VOC, we base our architecture on DeepLab~\citep{chen2017deeplab} with ResNet101~\citep{he2016deep} as the backbone network. For the experiments on DensePose, we adopt PSPNet~\citep{zhao2017pyramid} as the backbone network.  Our  models are pre-trained on ImageNet~\citep{deng2009imagenet} dataset.  
See Appendix for details on our inference procedure and hyper-parameter selection for training and testing.

For each type of annotations and dataset,
we formulate four types of pixel-to-segment contrastive relationships and jointly optimize them in a single pixel-wise metric learning framework (Fig.~\ref{fig:Main}).
Table~\ref{tab:annotations} shows the data-driven selection of hyperparameters $\lambdaImgSim,\lambdaSemCoc$ and $\lambdaFeatAff$ for different task settings.

\noindent \textbf{Pascal: Image tag annotations.}  
Table~\ref{tab:voc_image_box} shows that, 
without using additional saliency labels, our method  outperforms existing methods with saliency by $4.4\%$, and those without saliency by $5.1\%$.

\noindent \textbf{Pascal: Bounding box annotations.}
Table~\ref{tab:voc_image_box} shows that, with the same DeepLab/ResNet101 backbone network, our method outperforms existing methods by $3.2\%$.

\begin{table}[h]
\begin{minipage}{.48\linewidth}
\centering
\resizebox{\textwidth}{!}{%
\begin{tabular}{l|c|c|c}
    \Xhline{1pt}
    Pascal: Image tags & Saliency & {\it val} & {\it test}\\
    \hline
    ~\cite{huang2018weakly} & \checkmark & 61.4 & 63.2\\
    ~\cite{lee2019ficklenet} & \checkmark & 64.9 & 65.3\\
    ~\cite{zhang2019reliability} & - & 66.3 & 66.5\\
    ~\cite{yao2020saliency} & \checkmark & 67.1 & 67.2\\
    ~\cite{chang2020weak} &  - & 66.1 & 65.9\\
    \hline\hline
    Our \algorithmShort & - & \textbf{69.5} & \textbf{71.6}\\
    \Xhline{1pt}
\end{tabular}}
\end{minipage}%
\hfill
\begin{minipage}{.48\linewidth}
\centering
\begin{tabular}{l|c|c}
    \Xhline{1pt}
    Pascal: Bounding boxes & {\it val} & {\it test}\\
    \hline
    ~\cite{khoreva2017simple} & 69.4 & -\\
    ~\cite{song2019box} & 70.2 & -\\
    \hline\hline
    Our \algorithmShort & \textbf{73.5} & \textbf{74.7}\\
    \Xhline{1pt}
\end{tabular}
\end{minipage}
\vspace{-5pt}
\caption{Pascal VOC 2012 dataset with image tag (left) and bounding box (right) annotations.}
\label{tab:voc_image_box}
\end{table}

\begin{table}[h]
\begin{minipage}{0.5\linewidth}
\centering
\resizebox{\textwidth}{!}{%
\begin{tabular}{l|c|c|cc}
    \Xhline{1pt}
    Pascal: Scribbles & CRF & Full & Weak & WvF\\
    \hline
    ~\cite{tang2018normalized} &  & 75.6 & 72.8 & 96.3 \\
    ~\cite{tang2018normalized} & \checkmark & 76.8 & 74.5 & 97.0 \\
    ~\cite{tang2018regularized} &  & 75.6 & 73.0 & 96.6 \\
    ~\cite{tang2018regularized} & \checkmark & 76.8 & 75.0 & 97.7 \\
    ~\cite{wang2019boundary} &  & 75.6 & 73.2 & 96.8 \\
    ~\cite{wang2019boundary} & \checkmark & 76.8 & 76.0 & 99.0 \\
    \hline\hline
    Our \algorithmShort &  & 76.1 & \textbf{74.2} & \textbf{97.5} \\
    Our \algorithmShort & \checkmark & 77.3 & 76.1 & 98.4 \\
    \Xhline{1pt}
\end{tabular}}
\end{minipage}
\hfill
\begin{minipage}{0.5\linewidth}
    \includegraphics[width=0.9\textwidth]{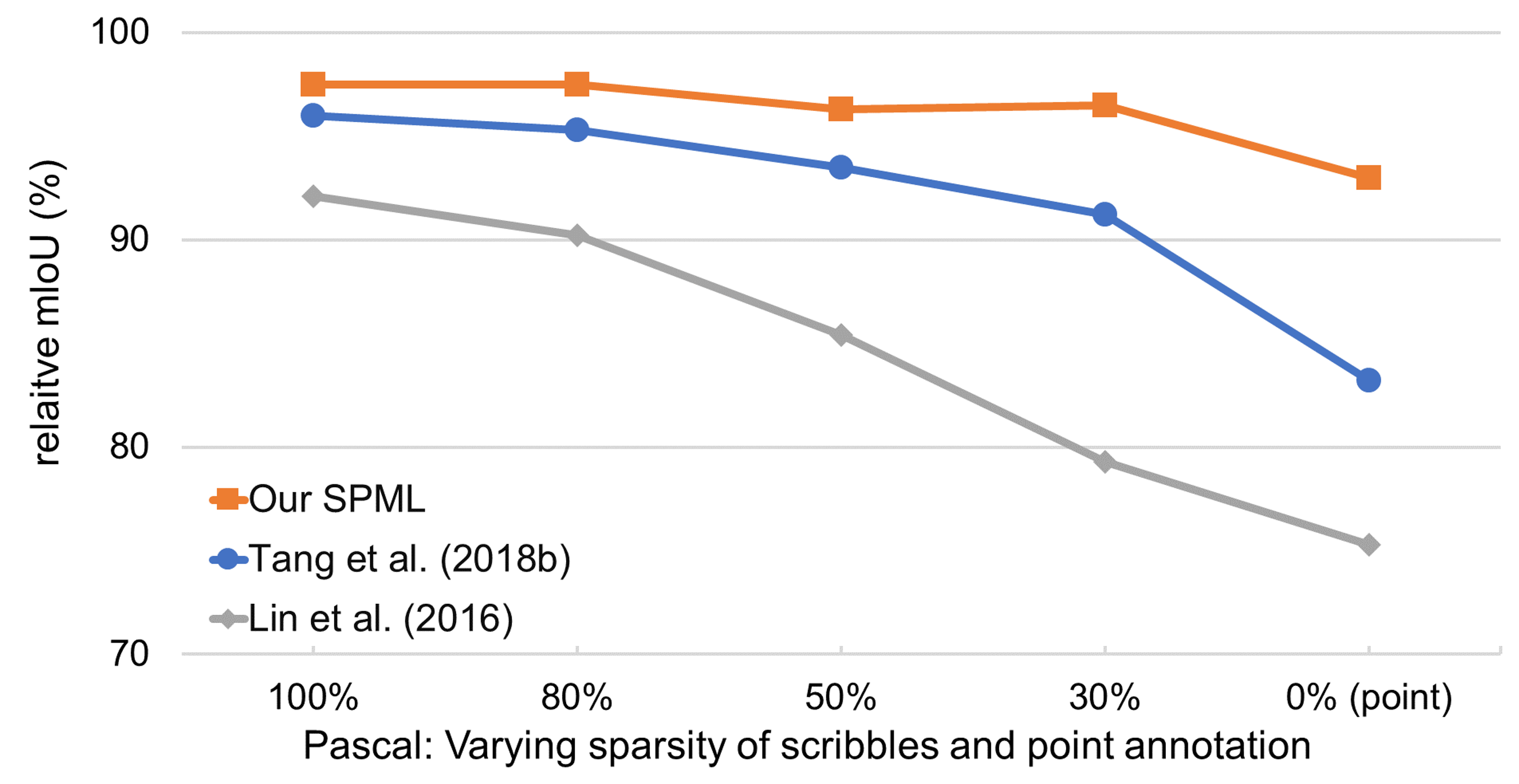}
\end{minipage}
\vspace{-5pt}
\caption{Pascal VOC 2012 dataset using scribble annotations. \textbf{Left}: mIoU on validataion (white) and test (gray) set. {\bf WvF} denotes relative mIoU w.r.t full supervision. \textbf{Right}: Relative mIoU performance w.r.t full supervision on different lengths of scribbles.}
\label{tab:voc_scribble}
\end{table}

\noindent \textbf{Pascal: Scribble annotations.} 
Table~\ref{tab:voc_scribble} shows that, 
our method consistently delivers the best performance among methods without or with CRF post-processing.  We get $74.2\%$ ($76.1\%$) mIoU, achieving $97.5\%$ ( $98.4\%$) of full supervision performance in these two categories respectively.

\noindent \textbf{Pascal: Varying sparsity of scribble and point annotations.}  Exploiting metric learning with different relationships in the data frees us from the classification framework and delivers a more powerful approach that requires fewer annotations. Table~\ref{tab:voc_scribble} shows that, as we shorten the length of scribbles from $100\%, 80\%, 50\%, 30\%$ to $0\%$ (points), we reach $97.5\%, 97.5\%, 96.3\%, 96.5\%$ and $93.7\%$ of full supervision performance.  Compared to the full scribble annotations, our accuracy only drops $3.7\%$ with point labels and is significantly better than the baseline.

\begin{figure}[!t]
    \centering
    \vspace{-20pt}
    \includegraphics[width=1.0\textwidth]{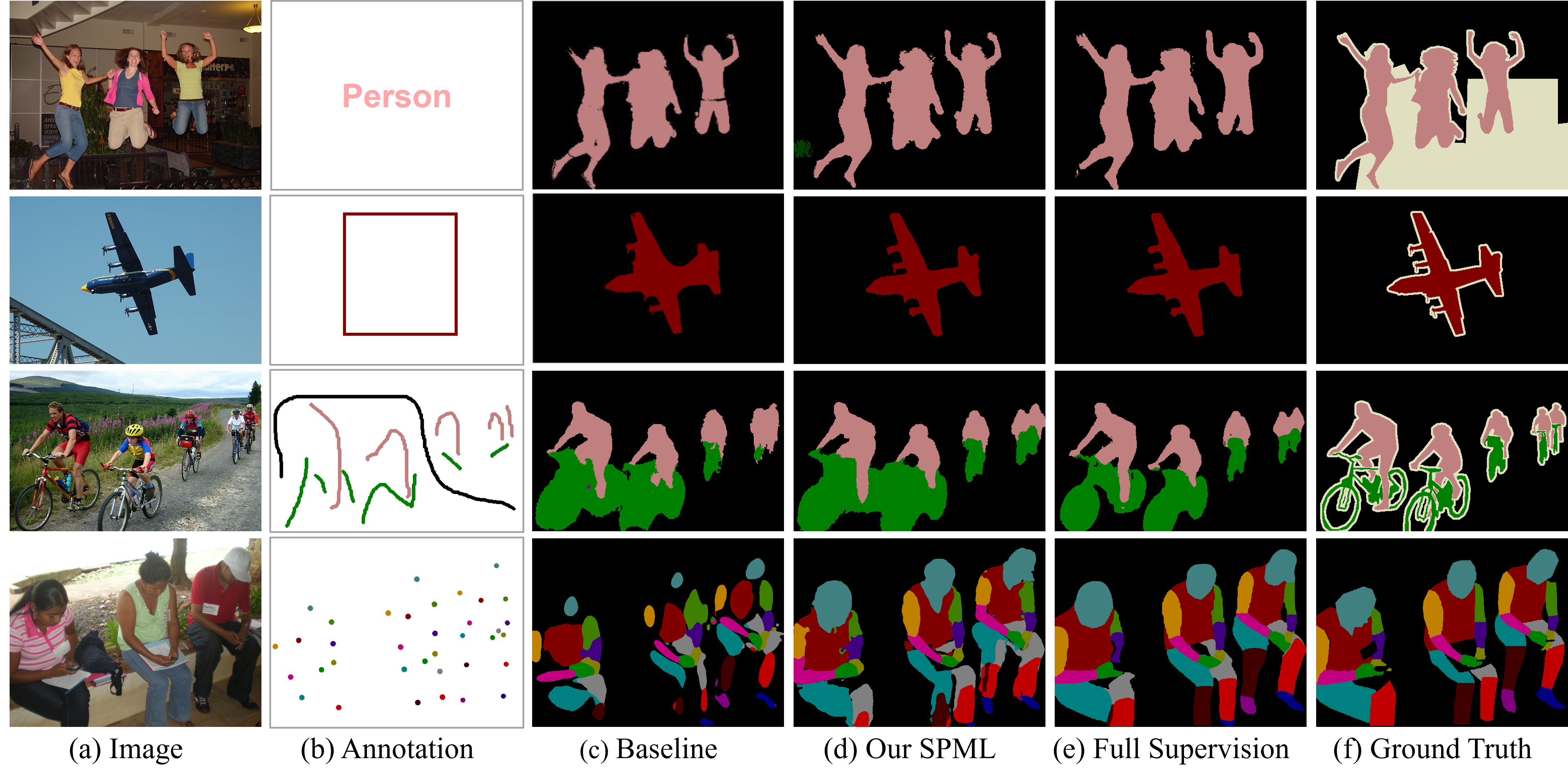}
    \vspace{-10pt}
    \caption{Our results on Pascal and DensePose  under various weak supervision settings are consistently better aligned with region boundaries and visually closer to fully supervised counterparts.}
    \label{fig:vis_results}
\end{figure}

\begin{figure}[t]
    \centering
    \includegraphics[width=\textwidth]{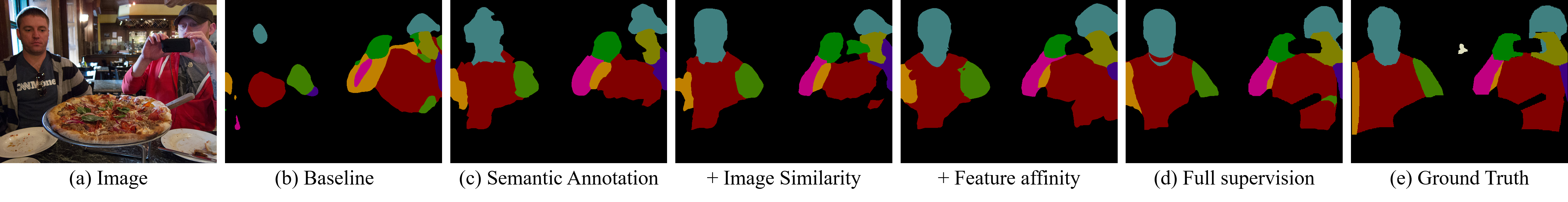}
    \vspace{-15pt}
    \caption{Our segmentation results get better with more types of regularizations. We compare visual results by adding more regularizations. As we introduce more relationships for regularization, we observe significant improvement and our results are visually closer to fully supervised counterparts.}
    \vspace{-5pt}
    \label{fig:vis_ablation_densepose}
\end{figure}

\begin{wraptable}{l}{55mm}
    \centering
    \begin{tabular}{@{}l|c|c@{}}
    \toprule
    DensePose: Points & mIoU & WvF\\
    \midrule
    ~\cite{tang2018regularized} & 31.3 & 51.9\\
    \midrule
    Our \algorithmShort & \textbf{44.2} & \textbf{77.1}\\
    \bottomrule
    \end{tabular}
    \vspace{-10pt}
    \caption{DensePose minival 2014 set.}
    \label{tab:densepose}
\end{wraptable}



\noindent\textbf{DensePose: Point annotations.} We train our baseline using the code released by \cite{tang2018regularized}.  Table~\ref{tab:densepose} shows that, our method without CRF post-processing
outperforms the baseline by $12.9\%$ mIoU, reaching $77.1\%$ of full supervision performance with only point supervision. 

\noindent\textbf{Visual quality and ablation study.}  Fig.~\ref{fig:vis_results} shows that our results are better aligned with region boundaries and visually closer to fully-supervised counterparts.  Fig.~\ref{fig:vis_ablation_densepose} shows that our results improve significantly with different relationships for more regularization.  See Appendix for more details and ablation studies.

%% file: 6conclusion.tex
\noindent
{\bf Summary.}
We propose a novel weakly-supervised semantic segmentation method via
Semi-supervised Pixel-wise Metric Learning, based on  four common types of pixel-to-segment attraction and repulsion relationships.  It is universally applicable to various weak supervision settings, whether the
training images are coarsely annotated by image tags or bounding boxes, or sparsely annotated by keypoints or scribbles.
Our results on PASCAL VOC and DensePose show consistent and substantial gains over SOTA, especially for the sparsest keypoint supervision.


%% file: 7appendix.tex
\def\figVisResults#1{
\begin{figure}[#1]
    \centering
    \includegraphics[width=\linewidth]{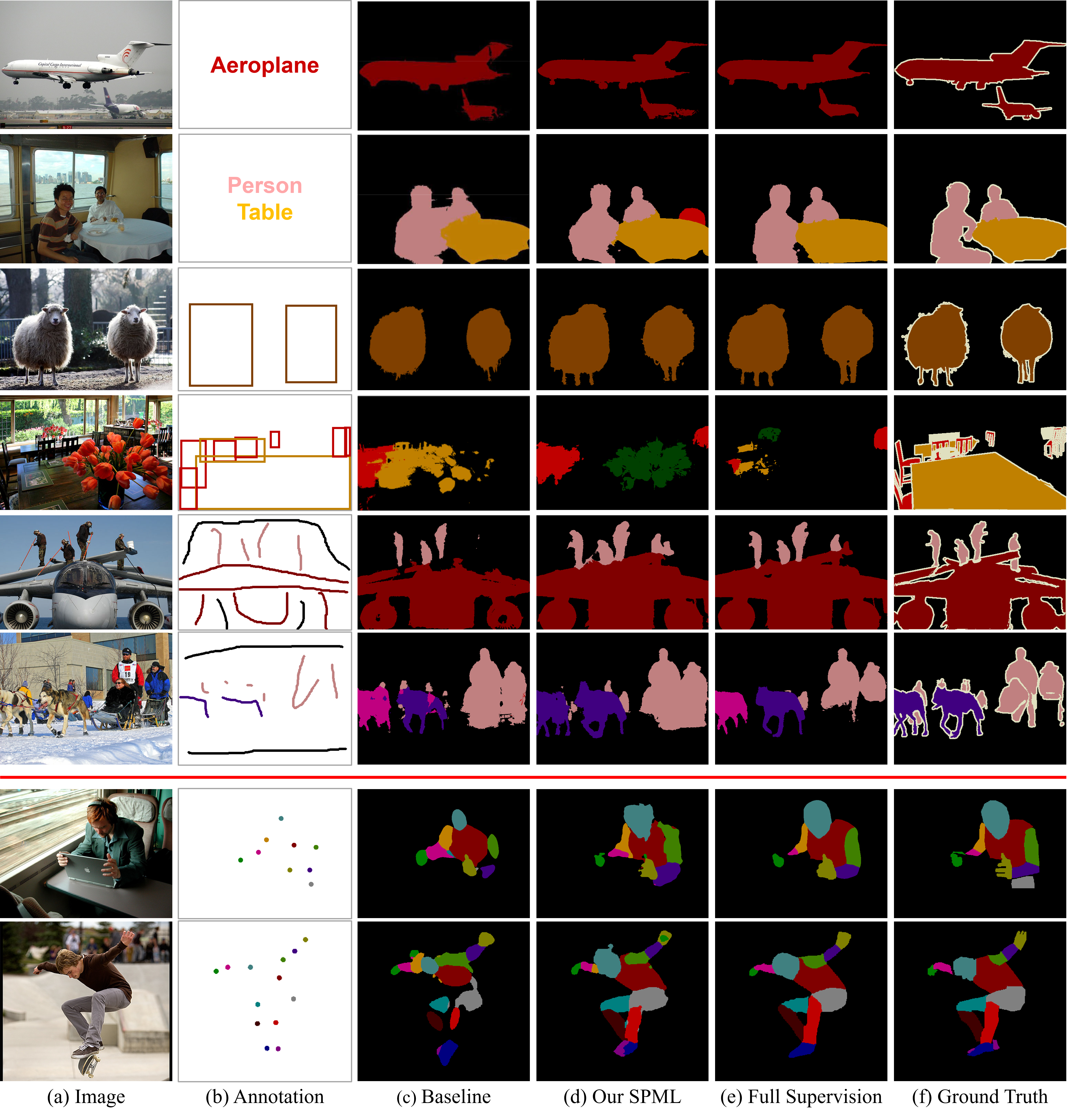}
    \caption{Visual comparison of baseline method (c), our SPML (d) and fully-supervised SegSort (e) on VOC and DensePose. On VOC (top 6 rows), our baseline method is based on ~\cite{lee2019ficklenet, song2019box, tang2018regularized} for image tag, bounding box and scribble annotations, respectively. On DensePose (bottom 2 rows), our baseline is ~\cite{tang2018regularized}. The results from our weakly-supervised model is visually very close to its fully-supervised counterpart, or even better when visual cues are prominent.}
    \label{fig:vis_results_appendix}
\end{figure}
}

\def\figVisAblation#1{
\begin{figure}[#1]
    \centering
    \includegraphics[width=\linewidth]{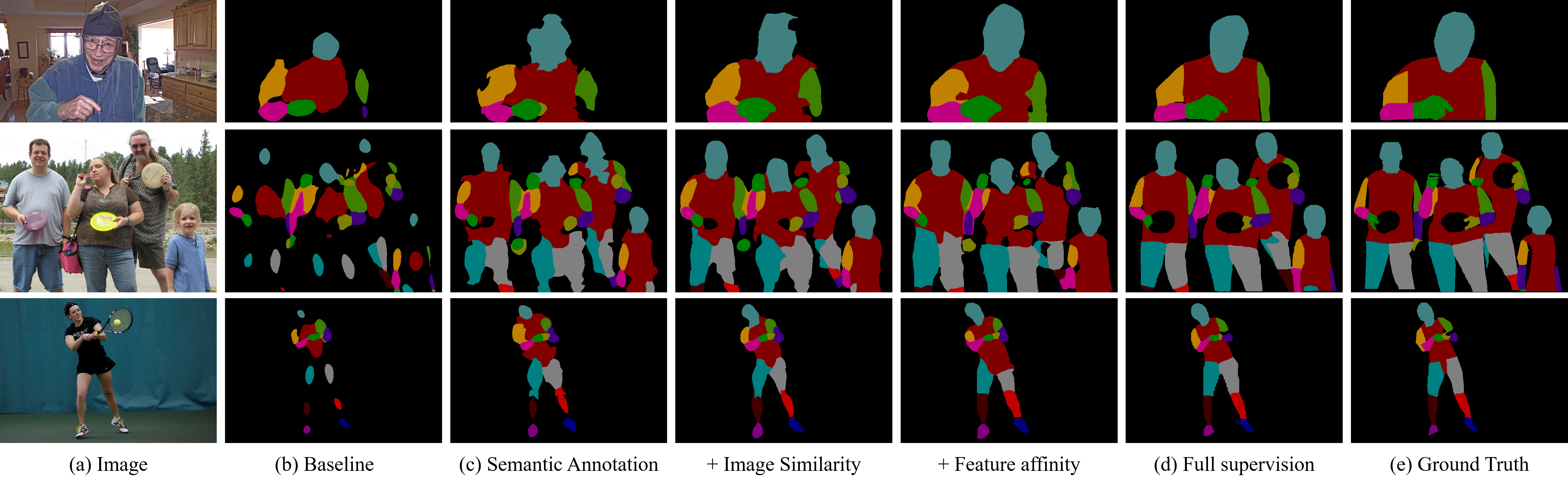}
    \caption{Our segmentation results get better with more types of regularizations. We compare visual results by adding more regularizations. As we introduce more relationships for regularization, we observe significant improvement and our results are visually closer to fully supervised counterparts.}
    \label{fig:vis_ablation_densepose_appendix}
\end{figure}
}

\def\figVisContext#1{
\begin{figure}[#1]
    \centering
    \includegraphics[width=\linewidth]{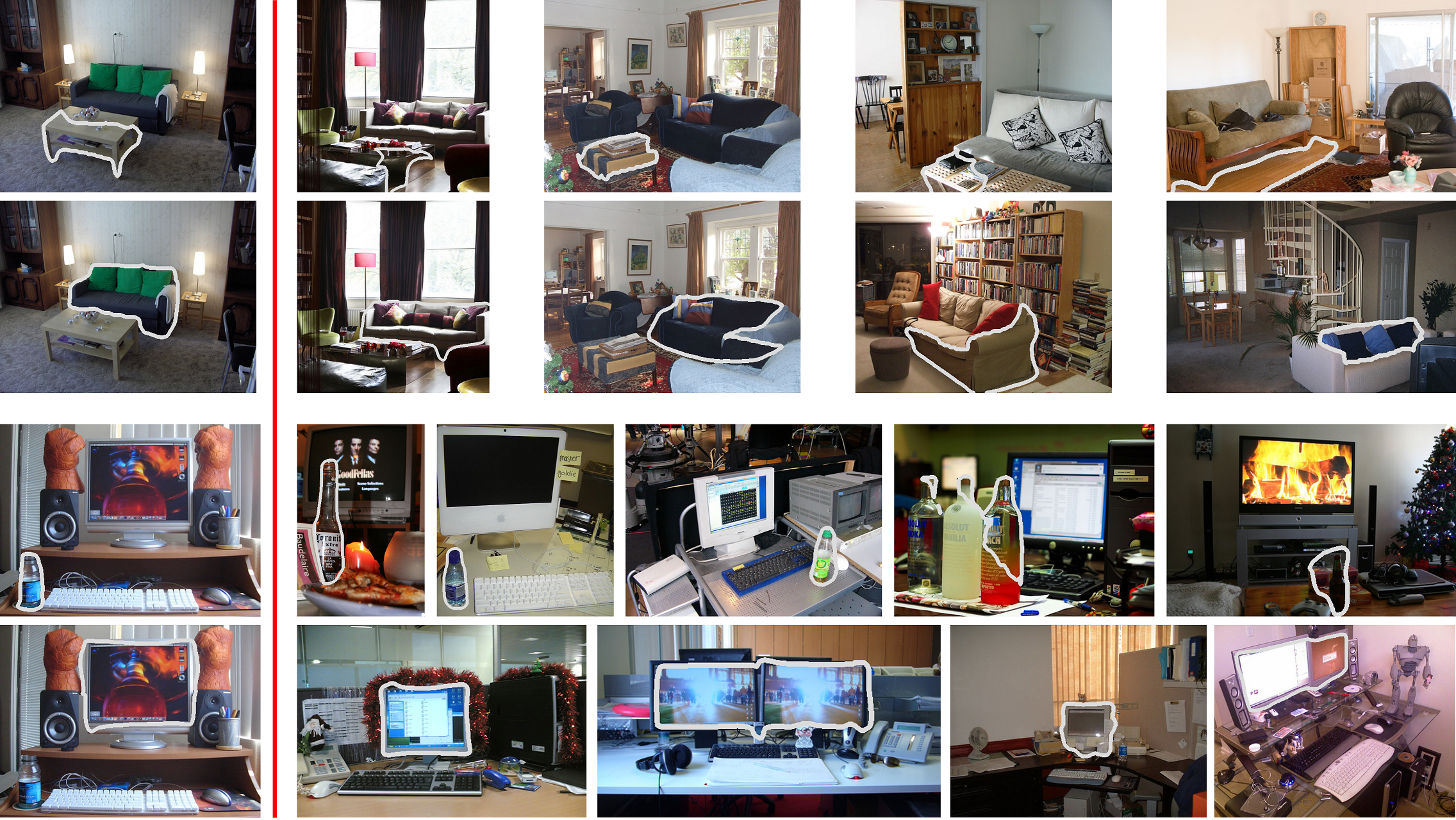}
    \caption{Visual examples of nearest neighbor segment retrievals. We observe that retrieved segments (right) appear in the similar semantic context as the query segments (left). For examples, given a bottle next to a desktop, our model retrieves bottles also next to a desktop. }
    \label{fig:vis_semantic_context}
\end{figure}
}

\def\figSemanticAnnotation#1{
\begin{figure}[#1]
    \centering
    \includegraphics[width=\linewidth]{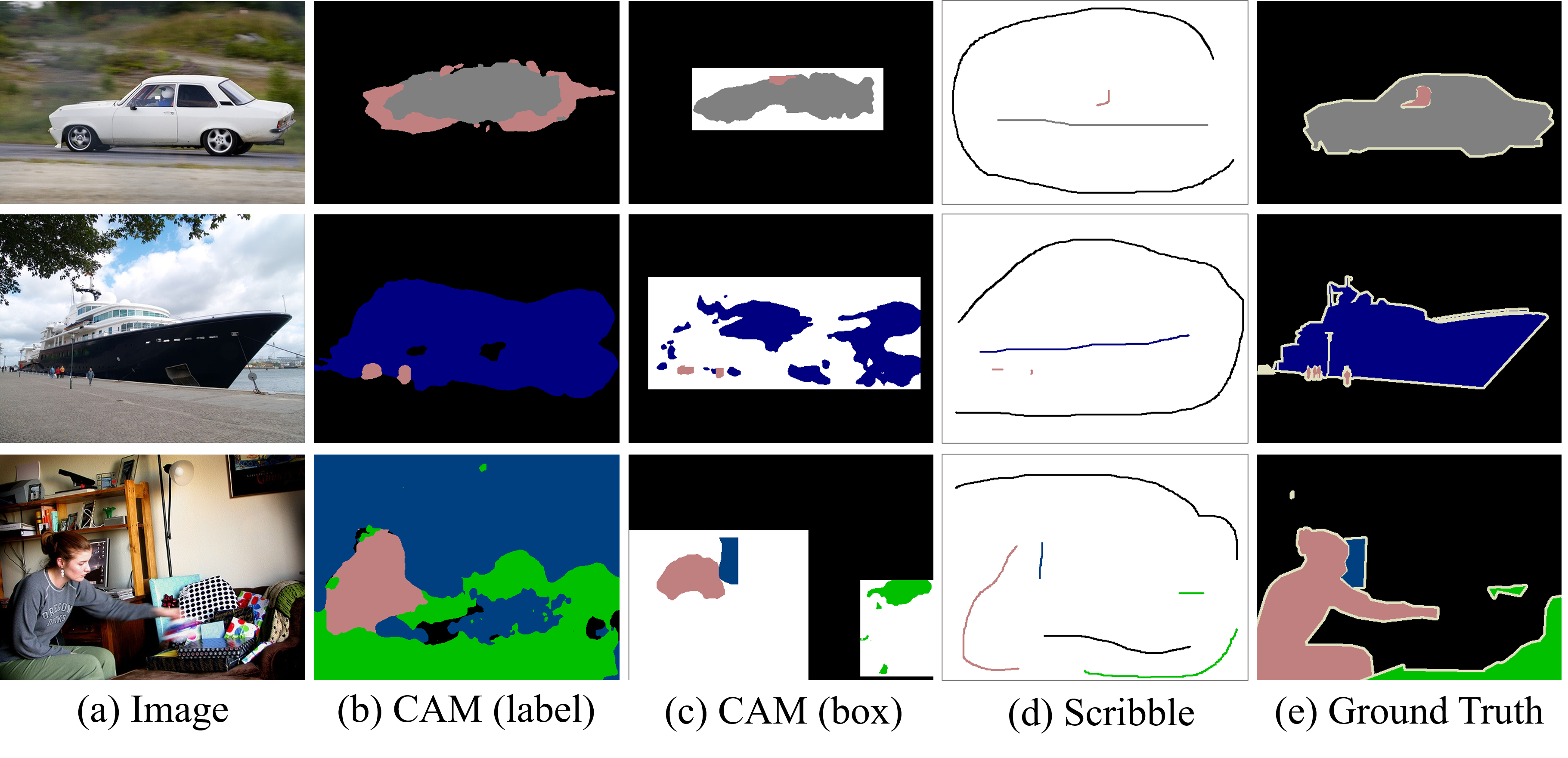}
    \caption{Visual examples of semantic annotations used on VOC. For image tag and bounding box annotation, we use the classifier trained by~\cite{wang2020self} to infer CAM as semantic annotation. These semantic annotations are noisy, which do not precisely localize on the objects.}
    \label{fig:vis_semantic_annotation}
\end{figure}
}

\def\figDenseposeProcess#1{
\begin{figure}[#1]
    \centering
    \includegraphics[width=\linewidth]{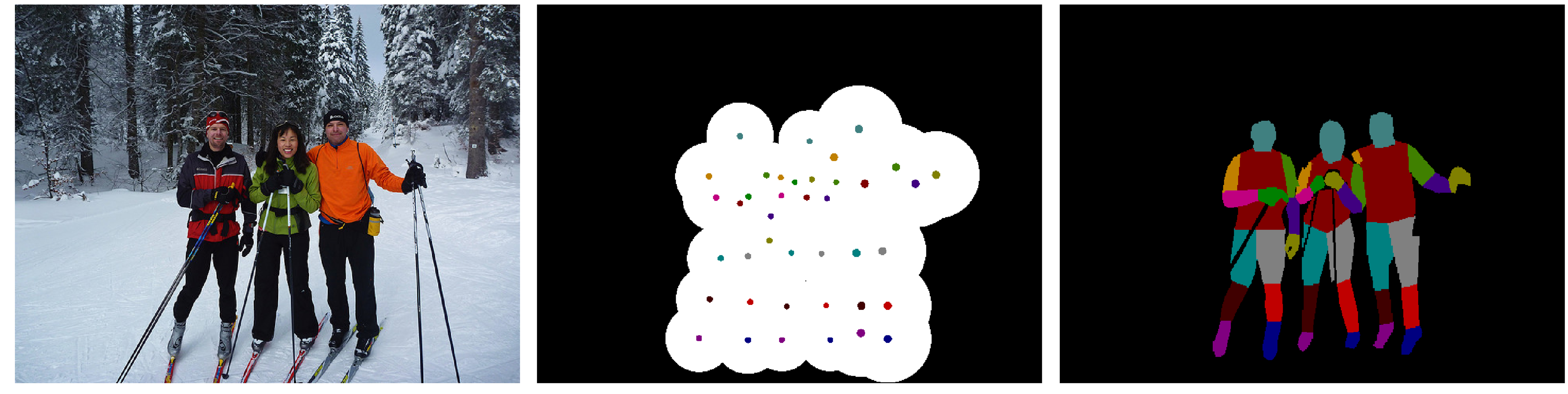}
    \caption{Preparing training labels on DensePose dataset. From left to right are input image, our training labels and ground-truth mask. For each keypoint, a Gaussian heat map is applied to determine labelled, unknown and background region. The white region denotes unknown pixels, to which we propagate labels from annotated or background region.}
    \label{fig:densepose_preprocess}
\end{figure}
}

\def\tabVocAblation#1{
    \begin{table}[#1]
    \parbox{.25\linewidth}{
    \centering
    \begin{tabular}{c|c|c}
    \toprule
    $\lambdaImgSim$ & $\lambdaSemCoc$ & mIoU\\
    \midrule
    0.3 & 0.5 & 73.7\\
    0.1 & 0.5 & 74.2 \\
    0.05 & 0.5 & 73.5\\
    0 & 0.5 & 71.7\\
    \bottomrule
    \end{tabular}}
    \hfill
    \parbox{.25\linewidth}{
    \centering
    \begin{tabular}{c|c|c}
    \toprule
    $\lambdaImgSim$ & $\lambdaSemCoc$ & mIoU\\
    \midrule
    0.1 & 1.0 & 74.1\\
    0.1 & 0.5 & 74.2 \\
    0.1 & 0.1 & 74.1\\
    0.1 & 0 & 72.8\\
    \bottomrule
    \end{tabular}}
    \hfill
    \parbox{.4\linewidth}{
    \centering
    \begin{tabular}{c|c|c|c}
    \toprule
    $\lambdaImgSim$ & $\lambdaSemCoc$ & $\lambdaFeatAff$ & mIoU\\
    \midrule
    0 & 0 & 0 & 71.2\\
    0.1 & 0 & 0 & 72.8\\
    0.1 & 0.5 & 0 & 74.2\\
    0.1 & 0.5 & 0.1 & 73.8\\
    \bottomrule
    \end{tabular}}
    \caption{Ablation study of different weighting parameters for each objective function on Pascal VOC validation dataset.}
    \label{tab:voc_ablation}
    \end{table}
}

\def\tabHyperParameters#1{
    \begin{table}[#1]
    \begin{center}
    \begin{tabular}{c|c|c c|c c|c c|c c|c}
        \toprule
        Dataset & Annotation & $\lambdaImgSim$ & $\concImgSim$ & $\lambdaSemAnn$ & $\concSemAnn$ & $\lambdaSemCoc$ & $\concSemCoc$ & $\lambdaFeatAff$ & $\concFeatAff$ & batchsize \\
        \midrule
        VOC & scribbles & 0.1 & 16 & 1.0 & 6 & 0.5 & 12 & 0.0 & - & 12\\
        & points & 1.0 & 16 & 1.0 & 6 & 1.0 & 8 & 0.0 & - & 12\\
        & boxes & 0.3 & 16 & 1.0 & 6 & 1.0 & 8 & 0.0 & - & 16\\
        & image tags & 0.3 & 16 & 1.0 & 6 & 1.0 & 8 & 0.0 & - & 16\\
        \midrule
        DensePose & points & 0.1 & 16 & 1.0 & 6 & 0.0 & - & 0.5 & 12 & 16\\
        \bottomrule
    \end{tabular}
    \end{center}
    \caption{Hyper-parameters for different types of annotations on Pascal and DensePose dataset.}
    \label{tab:hyperparams}
    \end{table}
}

\def\tabVocScribbleLength#1{
    \begin{table}[#1]
    \centering
    \resizebox{\textwidth}{!}{%
    \begin{tabular}{l|c|c|c|c c c c c}
    \Xhline{1pt}
    Method & Backbone & CRF & Full & 100\% & 80\% & 50\% & 30\% & 0\%\\
    \hline
    ~\cite{lin2016scribblesup} & DeepLab-MSc-LargeFOV & \checkmark & 68.5
     & 63.1 & 61.8 & 58.5 & 54.3 & 51.6\\
    ~\cite{tang2018regularized} & DeepLab-MSc-LargeFOV & \checkmark & 68.7 & 66.0 & 65.5 & 64.2 & 62.7 & 57.2\\
    \hline\hline
    Our \algorithmShort & DeepLab/ResNet101 & & 76.1 & 74.2 & 74.2 & 73.3 & 73.4 & 71.3\\
    Our \algorithmShort & DeepLab/ResNet101 & \checkmark & 77.3 & 76.1 & 75.8 & 74.8 & 75.0 & 73.2\\
    \Xhline{1pt}
    \end{tabular}}
    \caption{mIoU performance on Pascal VOC 2012 validation set on different lengths of scribble.}
    \label{tab:voc_scribble_len}
    \end{table}
}

\def\tabVocScribbleVal#1{
    \begin{table}[#1]
    \centering
    \resizebox{\linewidth}{!}{%
    \begin{tabular}{l|c c c c c c c c c c c c c c c c c c c c|c}
    \Xhline{1pt}
    Backbone & aero & bike & bird & boat & bottle & bus & car & cat & chair & cow & table & dog & horse & mbike & person & plant & sheep & sofa & train & tv & mIoU\\
    \hline\hline
    ~\cite{tang2018regularized} & 83.2 & 35.8 & 82.8 & 66.8 & 75.1 & 90.9 & 83.9 & 89.2 & 35.8 & 82.5 & 53.7 & 83.4 & 83.2 & 79.5 & 82.2 & 57.6 & 81.9 & 41.6 & 81.1 & 73.5 & 73.2\\
    Our \algorithmShort & \textbf{85.8} & \textbf{37.6} & 82.8 & \textbf{69.6} & \textbf{75.9} & 89.3 & 82.8 & \textbf{89.7} & \textbf{38.6} & \textbf{85.7} & \textbf{56.7} & \textbf{85.9} & 80.1 & 78.1 & \textbf{84.8} & 53.9 & \textbf{83.7} & \textbf{49.2} & 80.9 & \textbf{74.4} & \textbf{74.2}\\
    \hline\hline
    \rowcolor{Gray}
    ~\cite{tang2018regularized} & 86.2 & 37.3 & 85.5 & 69.4 & 77.8 & 91.7 & 85.1 & 91.2 & 38.8 & 85.1 & 55.5 & 85.6 & 85.8 & 81.7 & 84.1 & 61.4 & 84.3 & 43.1 & 81.4 & 74.2 & 75.2\\
    \rowcolor{Gray}
    Our \algorithmShort & \textbf{89.0} & \textbf{38.4} & \textbf{86.0} & \textbf{72.6} & \textbf{77.9} & 90.0 & 83.9 & 91.0 & \textbf{40.0} & \textbf{88.3} & \textbf{57.7} & \textbf{87.7} & 82.8 & 79.1 & \textbf{86.5} & 57.1 & \textbf{87.4} & \textbf{50.5} & 81.2 & \textbf{76.9} & \textbf{76.1}\\
    \Xhline{1pt}
    \end{tabular}}
    \caption{Per-class results on Pascal VOC 2012 validation set. White- and gray-colored background denotes using without- and with- CRF post-processing for inference.}
    \label{tab:voc_scribble_val_appendix}
    \end{table}
}

\def\tabVocScribbleTest#1{
    \begin{table}[#1]
    \centering
    \resizebox{\linewidth}{!}{%
    \begin{tabular}{l|c c c c c c c c c c c c c c c c c c c c|c}
    \Xhline{1pt}
    Annotations & aero & bike & bird & boat & bottle & bus & car & cat & chair & cow & table & dog & horse & mbike & person & plant & sheep & sofa & train & tv & mIoU\\
    \hline\hline
    Full mask & 91.5 & 43.5 & 83.0 & 67.9 & 81.7 & 89.8 & 88.7 & 94.6 & 37.5 & 81.6 & 68.7 & 88.8 & 82.4 & 88.6 & 87.6 & 64.1 & 87.6 & 52.7 & 76.5 & 71.4 & 77.3\\
    \hline\hline
    Scribbles & 87.0 & 36.7 & 82.3 & 65.5 & 79.7 & 89.5 & 84.8 & 90.1 & 37.6 & 86.3 & 63.1 & 89.1 & 87.8 & 83.0 & 86.0 & 65.8 & 85.8 & 60.3 & 76.9 & 73.0 & 76.4\\
    Points & 83.5 & 37.0 & 78.4 & 61.9 & 74.8 & 86.4 & 83.2 & 86.9 & 37.9 & 85.3 & 62.4 & 87.2 & 84.2 & 81.1 & 83.1 & 64.3 & 85.1 & 59.1 & 74.0 & 66.3 & 74.0\\
    Boxes & 84.1 & 36.5 & 86.7 & 57.6 & 75.7 & 87.7 & 84.8 & 89.6 & 39.4 & 86.4 & 57.2 & 89.2 & 88.0 & 82.6 & 80.3 & 54.7 & 88.2 & 55.9 & 79.7 & 71.6 & 74.7\\
    Tags & 82.1 & 38.7 & 80.0 & 56.9 & 73.7 & 85.7 & 81.0 & 86.7 & 33.9 & 87.7 & 60.8 & 86.8 & 84.9 & 81.3 & 77.7 & 53.2 & 86.5 & 50.1 & 64.8 & 58.4 & 71.6\\
    \Xhline{1pt}
    \end{tabular}}
    \caption{Per-class results on Pascal VOC 2012 testing set. CRF post-processing is used for inference.}
    \label{tab:voc_scribble_test_appendix}
    \end{table}
}

\def\tabDenseposeVal#1{
    \begin{table}[#1]
        \centering
        \resizebox{\textwidth}{!}{%
        \begin{tabular}{l|c c c c c c c c c c c c c c c|c|c}
        \Xhline{1pt}
        Method & bg. & torso & RHand & LHand & LFoot & RFoot & RThigh & LThigh & RLeg & LLeg & LArm & RArm & LFarm & RFarm & Heaad & mIoU & WvF\\
        \hline \hline
        Softmax & 96.2 & 73.7 & 61.1 & 57.2 & 37.2 & 37.8 & 56.8 & 54.8 & 49.7 & 49.5 & 62.0 & 63.8 & 58.3 & 61.5 & 84.6 & 60.3 & -\\
        SegSort & 95.8 & 71.9 & 57.4 & 53.0 & 33.4 & 33.4 & 54.0 & 51.8 & 46.4 & 46.9 & 59.2 & 61.1 & 54.4 & 57.9 & 83.2 & 57.3 & -\\
        \hline\hline
        \rowcolor{Gray}
        ~\cite{tang2018regularized} & 87.2 & 28.3 & 37.5 & 36.0 & 18.9 & 19.5 & 21.2 & 20.8 & 16.1 & 16.6 & 33.9 & 35.3 & 35.6 & 37.6 & 25.2 & 31.3 & 51.9\\
        \rowcolor{Gray}
        Our \algorithmShort & 93.8 & 57.7 & 48.1 & 43.2 & 22.8 & 22.2 & 36.6 & 35.6 & 27.1 & 27.6 & 42.1 & 45.3 & 42.0 & 45.5 & 72.6 & 44.2 & 77.1\\
        \Xhline{1pt}
        \end{tabular}}
        \caption{Per-class results on DensePose minival 2014 set with keypoint annotations. White- and gray-colored background indicates using full and point supervision.}
        \label{tab:densepose_appendix}
    \end{table}
}

\def\algTagInference#1{
    \begin{algorithm}[#1]
    \SetAlgoLined
    \SetNoFillComment
    \LinesNumbered
    {\bf Input:} Fixed pixel-wise embedding $\pmb{e}$ of the input image and CAM logits $\mathcal{M}$.
    
    {\bf Output:} Semantic segmentation prediction $\mathcal{Y}_{pred}$.
    
    \tcc{Train the initial softmax classifier}
    Calculate pixel-wise transition probability matrix $T$ from $\pmb{e}$.
    
    Refine CAM by random walk propagation: $\mathcal{M}' = T^\top \circ ... \circ T^\top \mathcal{M}$.
    
    Derive pseudo labels from refined CAM: $\mathcal{Y}^1_{cam}=\argmax_c \mathcal{M}'_c$.
    
    Predict new pseudo labels $\mathcal{Y}^1_{nn}$ from $\mathcal{Y}^1_{cam}$ using nearest neighbor retrievals.
    
    Train the softmax classifier $S_1$ using $\mathcal{Y}^1_{nn}$.
    
    \tcc{Train the final softmax classifier}
    
    Predict pseudo labels $\mathcal{Y}^2_{sc}$ from initial softmax classifier $S_1$.
    
    Predict new pseudo labels $\mathcal{Y}^2_{nn}$ from $\mathcal{Y}^2_{sc}$ using nearest neighbor retrievals.
    
    Train the softmax classifier $S_2$ using $\mathcal{Y}^2_{nn}$.
    
    Predict final semantic segmentation $\mathcal{Y}_{pred}$ from $S_2$.
   
    \caption{Inference procedure for semantic segmentation using image-level tags.}
    \label{alg:tag_inference}
    \end{algorithm}
}

\def\algInference#1{
    \begin{algorithm}[#1]
    \SetAlgoLined
    \SetNoFillComment
    \LinesNumbered
    {\bf Input:} Fixed pixel-wise embedding $\pmb{e}$ of the input image and weak annotations $\mathcal{Y}_{weak}$.
    
    {\bf Output:} Semantic segmentation prediction $\mathcal{Y}_{pred}$.
    
    \tcc{Train the initial softmax classifier}
    
    Train the softmax classifier $S_1$ using $\mathcal{Y}_{weak}$.
    
    \tcc{Train the final softmax classifier}
    
    Predict semantic logits from initial softmax classifier: $\mathcal{\tilde{M}}=S_1(\pmb{e})$.
    
    Calculate pixel-wise transition probability matrix $T$ from $\pmb{e}$.
    
    Refine semantic logits by random walk propagation: $\mathcal{\tilde{M}}' = T^\top \circ ... \circ T^\top  \mathcal{\tilde{M}}$.
    
    Derive pseudo labels from refined semantic logits: $\mathcal{Y}_{sc}=\argmax_c \mathcal{\tilde{M}}'_c$.

    Train the softmax classifier $S_2$ using $\mathcal{Y}_{sc}$.
    
    Predict final semantic segmentation $\mathcal{Y}_{pred}$ from $S_2$.
   
    \caption{Inference procedure for semantic segmentation using scribble / point / bounding box annotations.}
    \label{alg:inference}
    \end{algorithm}
}

\appendix
\section{Appendix}

We propose a single pixel-to-segment contrastive learning loss formulation for weakly supervised semantic segmentation. We explore different types of visual relationships to group and separate pixels {\it within} and {\it across} images. We demonstrate state-of-the-art performance using our proposed method with different types of annotations. Here, we include more details on the following aspects:

\begin{itemize}
    \item We present the visual results of our method in ~\ref{supsec:vis}.
    
    \item We showcase the semantic cues generated by CAM for image tag and bounding box annotations in ~\ref{supsec:annotation}.
 
     \item We illustrate the data pre-processing used for DensePose dataset in ~\ref{supsec:densepose_preprocess}.

    \item We describe the details of our experimental settings, hyper-parameters and inference procedure in ~\ref{supsec:exp_setup}.
    
    \item We present the ablation study regarding hyper-parameters in ~\ref{supsec:ablation_study}.
    
    \item We present mIoU performance with varying sparsity of scribble annotations on Pascal dataset in ~\ref{supsec:scribble_len}.
    
    \item We present per-category results with Pascal and DensePose dataset in ~\ref{supsec:percat_miou}.
    

\end{itemize}

\figVisResults{b!}
\figVisAblation{t!}
\figVisContext{b!}

\subsection{Visualization}
\label{supsec:vis}
We present the visual results on VOC (with image tags, bounding boxes and scribbles) and DensePose (with keypoints) dataset in figure~\ref{fig:vis_results_appendix}. We observe that our segmentation results are better aligned with image boundary. When visual evidence is prominent, our weakly-supervised results are even better than the fully-supervised counterpart.

We then demonstrate the efficacy of each visual relationship in figure~\ref{fig:vis_ablation_densepose_appendix}. By adding {\bf semantic annotation}, {\bf low-level image similarity} and {\bf feature affinity} progressively, we observe consistent improvement of our results. The predicted segmentation becomes more coherent and better aligned with image boundary.
We lastly showcase that our method implicitly encodes semantic contexts. In figure~\ref{fig:vis_semantic_context}, We observe that retrieved segments appear in the similar semantic context as the query segments. For examples, given a bottle next to a desktop, our model retrieves bottles also next to a desktop; a set of sofas in a living room can be retrieved using one sofa query example; screens of a desktop can also be retrieved likewise.

\figSemanticAnnotation{t!}

\subsection{Semantic Annotations}
\label{supsec:annotation}
Since image tag and bounding box annotations do not provide any of precisely localized semantic information, we adopt CAM~\citep{zhou2016learning} to produce localized semantic cues. Without using additional saliency labels, we use the classifier trained by~\cite{wang2020self} to generate CAM. Let $\mathcal{M}_c$ be the activation map of class $c$. 

For image tag annotations, we follow ~\cite{ahn2018learning} to normalize $\mathcal{M}_c$ of the entire image within the range between 0 and 1, where $\mathcal{M}_c = \frac{\mathcal{M}_c}{\max_{c}{\mathcal{M}_c}}$. The background confidence $\mathcal{M}_{bg}$ can then be estimated by $\mathcal{M}_{bg} = (1 - \max_{c}\mathcal{M}_c)^\alpha$, where $\alpha$ is the hyper-parameter adjusting background confidence. In our experiments, we set $\alpha$ to $6$ and confidence threshold to $0.2$. The low-confidence pixels are considered as unlabeled regions.

For bounding box annotations, we simply normalize the CAM logits within each bounding box to the range between 0 and 1. We then set confidence threshold to $0.5$ for selecting foreground pixels and unlabeled regions. We restrict all the regions outside bounding boxes as ``background''. See figure~\ref{fig:vis_semantic_annotation} for more visual examples.

\figDenseposeProcess{b!}
\subsection{Data pre-processing for DensePose dataset}
\label{supsec:densepose_preprocess}
We next illustrate our pre-processing to generate training labels given keypoint annotations in DensePose dataset. As shown in figure~\ref{fig:densepose_preprocess}, we first assume a Gaussian heat map from every keypoint. By thresholding, we derive 3 regions from every Gaussian blob: labelled, unknown and background region. In labelled region, pixels are annotated as each body part. We then propagate labels, including background class, to pixels in the unknown region. The $std$ of Gaussian heat map is estimated from instance size, and we use ground-truth information in our paper.

\subsection{Hyper-parameters and Experimental Setup}
\label{supsec:exp_setup}

\noindent \textbf{Architecture and training.}
For all the experiments on VOC, we base our architecture as DeepLab~\citep{chen2017deeplab} with ResNet101~\citep{he2016deep} as backbone network. For the experiments on DensePose dataset, we adopt PSPNet~\citep{zhao2017pyramid} as backbone network. We only use models pre-trained on ImageNet~\citep{deng2009imagenet} dataset. 

We next describe the hyper-parameters used for each experiment. On Pascal VOC dataset, we set ``batchsize'' to $12$ and $16$ for scribble / point and image tag / bounding box annotations. On DensePose dataset, ``batchsize'' is set to $16$. For all the experiments, we train our models with $512 \times 512$ ``cropsize''. Following ~\cite{chen2017deeplab}, we adopt poly learning rate policy by multiplying base learning rate by $1 - (\frac{iter}{max\_iter})^{0.9}$. We set initial learning rate to $0.003$, momentum to $0.9$. For the hyper-parameters in SegSort framework, we use unit-length normalized embedding of dimension $64$ and $32$ on VOC and DensePose, respectively. We iterate K-Means clustering for $10$ iterations and generate $36$ and $144$ clusters on VOC and DensePose dataset. We set the concentration parameter $\kappa$ to different values for {\bf semantic annotation}, {\bf low-level image similarity}, {\bf semantic co-occurrence} and {\bf feature affinity}, respectively. Moreover, $\lambdaImgSim,\lambdaSemCoc$ and $\lambdaFeatAff$ are set to different values according to different types of annotations and datasets. $\lambdaSemAnn$ is set to 1 among all the experiments. The detailed hyper-parameter settings are summarized in table~\ref{tab:hyperparams}. We train for $30k$ and $45k$ iterations on VOC and DensePose dataset for all the experiments. We use additional memory banks to cache up previous 2 batches. For conducting experiments, we take advantage of XSEDE infrastructure~\citep{towns2014xsede} that includes Bridges resources~\citep{nystrom2015bridges}.

\noindent \textbf{Inference and testing.}
We fix the learned pixel-wise embedding and train an additional softmax classifier for inference. Iterative training is adopted to bootstrap the semantic segmentation prediction. Notably, we do not propagate gradients to the segmentation CNN from the softmax classifier.

For scribbles / points / bounding boxes, we first learn an initial softmax classifier $S_1$ from the corresponding weak annotations. Following ~\cite{ahn2018learning}, we apply random walk to refine the semantic logits $\mathcal{\tilde{M}}$ generated by $S_1$. The transition probability matrix $T$ is formulated as follows: $T_{i,j} = (\frac{\exp(\gamma \pmb{e}_i^\top \pmb{e}_j)}{\sum_j \exp(\gamma \pmb{e}_i^\top \pmb{e}_j)})^\beta$, where $\beta$ and $\gamma$ are 20 and 5, respectively. The label propagation is given by: $\mathcal{\tilde{M}}'=T^\top \mathcal{\tilde{M}}$, where $\mathcal{\tilde{M}}'$ denotes refined semantic logits. The random walk process is iterated for 6 times. Next, we obtain the corresponding pseudo labels $\mathcal{Y}_{sc}=\argmax_c \mathcal{\tilde{M}}_c'$. The pseudo labels are used to train the final softmax classifier $S_2$ for predicting semantic segmentation.

For image tag annotations, we adopt both within-image and across-image label propagation to generate optimal pseudo labels. Starting with CAM logits $\mathcal{M}$, we conduct within-image label propagation thru random walk and obtain refined pseudo labels $\mathcal{Y}^1_{cam}$. Across-image label propagation is carried out by nearest neighbor search thru the whole training set. We refer to SegSort~\citep{hwang2019segsort} for more details. We then obtain refined pseudo labels $\mathcal{Y}^1_{nn}$ and train the initial softmax classifier $S_1$. Similarly, we use $S_1$ to predict pseudo labels $\mathcal{Y}^2_{sc}$ from the training images. Followed by nearest neighbor search, we obtain our final pseudo labels $\mathcal{Y}^2_{nn}$ and train the final semantic classifier $S_2$.

The inference procedures for different annotations are summarized in algorithm~\ref{alg:inference} and ~\ref{alg:tag_inference}, respectively. For image tags, we adopt multi-scale and horizontally flipping as data augmentation for predicting semantic segmentation. For scribbles / points / bounding boxes, we do not employ data augmentation during the final inference.

\tabHyperParameters{h}
\algInference{h}
\algTagInference{h}

\tabVocAblation{h}

%
\subsection{Ablation study of hyper-parameters}
\label{supsec:ablation_study}
We conduct ablation study over different regularizations on Pascal VOC dataset. As shown in table~\ref{tab:voc_ablation}, we achieve the most optimal performance on Pascal VOC dataset with $\lambdaImgSim=0.1$ and $\lambdaSemCoc=0.5$. We also observe performance drops $0.4$ of mIoU by adding {\bf feature affinity} regularization. We argue that scribble/box/point annotations are not uniformly distributed across object instance and background, and results in noisy label propagation. 

\tabVocScribbleLength{h}

\tabVocScribbleVal{h!}

\tabVocScribbleTest{h!}

\tabDenseposeVal{b!}


\subsection{mean IoU performance with varying sparsity of scribbles.}
\label{supsec:scribble_len}
We report absolute mIoU performance by varying sparsity of scribbles on Pascal VOC 2012 validation set. The results are summarized in table~\ref{tab:voc_scribble_len}. Our results are much better with sparser annotation.

\subsection{Per-category mIoU on Pascal VOC and DensePose dataset.}
\label{supsec:percat_miou}
We next present per-category results on Pascal VOC and Denspose dataset. In table~\ref{tab:voc_scribble_val_appendix}, we compare with ~\cite{tang2018regularized} on VOC validation set. Without- and with CRF post-processing, our method outperform the baseline method among most categories by large margin. We further conduct experiments on VOC testing set, using DeepLab as backbone network. In table~\ref{tab:voc_scribble_test_appendix}, we can retrieve most performance w.r.t full supervision. We also compare per-category results on DensePose dataset in table~\ref{tab:densepose_appendix}. We train our baseline method using the code released by ~\cite{tang2018regularized}. We outperform the baseline method by large margin in every category.

%% file: 0main.bbl
\begin{thebibliography}{55}
\providecommand{\natexlab}[1]{#1}
\providecommand{\url}[1]{\texttt{#1}}
\expandafter\ifx\csname urlstyle\endcsname\relax
  \providecommand{\doi}[1]{doi: #1}\else
  \providecommand{\doi}{doi: \begingroup \urlstyle{rm}\Url}\fi

\bibitem[Ahn \& Kwak(2018)Ahn and Kwak]{ahn2018learning}
Jiwoon Ahn and Suha Kwak.
\newblock Learning pixel-level semantic affinity with image-level supervision
  for weakly supervised semantic segmentation.
\newblock In \emph{Proceedings of the IEEE Conference on Computer Vision and
  Pattern Recognition}, pp.\  4981--4990, 2018.

\bibitem[Alp~G{\"u}ler et~al.(2018)Alp~G{\"u}ler, Neverova, and
  Kokkinos]{alp2018densepose}
R{\i}za Alp~G{\"u}ler, Natalia Neverova, and Iasonas Kokkinos.
\newblock Densepose: Dense human pose estimation in the wild.
\newblock In \emph{Proceedings of the IEEE Conference on Computer Vision and
  Pattern Recognition}, pp.\  7297--7306, 2018.

\bibitem[Araslanov \& Roth(2020)Araslanov and Roth]{araslanov2020single}
Nikita Araslanov and Stefan Roth.
\newblock Single-stage semantic segmentation from image labels.
\newblock \emph{Proceedings of the IEEE conference on computer vision and
  pattern recognition}, 2020.

\bibitem[Arbelaez et~al.(2010)Arbelaez, Maire, Fowlkes, and
  Malik]{arbelaez2010contour}
Pablo Arbelaez, Michael Maire, Charless Fowlkes, and Jitendra Malik.
\newblock Contour detection and hierarchical image segmentation.
\newblock \emph{IEEE transactions on pattern analysis and machine
  intelligence}, 33\penalty0 (5):\penalty0 898--916, 2010.

\bibitem[Banerjee et~al.(2005)Banerjee, Dhillon, Ghosh, and
  Sra]{banerjee2005clustering}
Arindam Banerjee, Inderjit~S Dhillon, Joydeep Ghosh, and Suvrit Sra.
\newblock Clustering on the unit hypersphere using von mises-fisher
  distributions.
\newblock \emph{Journal of Machine Learning Research}, 2005.

\bibitem[Bearman et~al.(2016)Bearman, Russakovsky, Ferrari, and
  Fei-Fei]{bearman2016s}
Amy Bearman, Olga Russakovsky, Vittorio Ferrari, and Li~Fei-Fei.
\newblock What’s the point: Semantic segmentation with point supervision.
\newblock In \emph{European conference on computer vision}, pp.\  549--565.
  Springer, 2016.

\bibitem[Chang et~al.(2020)Chang, Wang, Hung, Piramuthu, Tsai, and
  Yang]{chang2020weak}
Yu-Ting Chang, Qiaosong Wang, Wei-Chih Hung, Robinson Piramuthu, Yi-Hsuan Tsai,
  and Ming-Hsuan Yang.
\newblock Weakly-supervised semantic segmentation via sub-category exploration.
\newblock \emph{Proceedings of the IEEE conference on computer vision and
  pattern recognition}, 2020.

\bibitem[Chen et~al.(2017)Chen, Papandreou, Kokkinos, Murphy, and
  Yuille]{chen2017deeplab}
Liang-Chieh Chen, George Papandreou, Iasonas Kokkinos, Kevin Murphy, and Alan~L
  Yuille.
\newblock Deeplab: Semantic image segmentation with deep convolutional nets,
  atrous convolution, and fully connected crfs.
\newblock \emph{IEEE transactions on pattern analysis and machine
  intelligence}, 40\penalty0 (4):\penalty0 834--848, 2017.

\bibitem[Clark et~al.(2018)Clark, Luong, and Le]{clark2018cross}
Kevin Clark, Thang Luong, and Quoc~V Le.
\newblock Cross-view training for semi-supervised learning.
\newblock \emph{arXiv preprint arXiv:1809.08370}, 2018.

\bibitem[Dai et~al.(2015)Dai, He, and Sun]{dai2015boxsup}
Jifeng Dai, Kaiming He, and Jian Sun.
\newblock Boxsup: Exploiting bounding boxes to supervise convolutional networks
  for semantic segmentation.
\newblock In \emph{Proceedings of the IEEE International Conference on Computer
  Vision}, pp.\  1635--1643, 2015.

\bibitem[Deng et~al.(2009)Deng, Dong, Socher, Li, Li, and
  Fei-Fei]{deng2009imagenet}
Jia Deng, Wei Dong, Richard Socher, Li-Jia Li, Kai Li, and Li~Fei-Fei.
\newblock Imagenet: A large-scale hierarchical image database.
\newblock In \emph{2009 IEEE conference on computer vision and pattern
  recognition}, pp.\  248--255. Ieee, 2009.

\bibitem[Everingham et~al.(2010)Everingham, Van~Gool, Williams, Winn, and
  Zisserman]{everingham2010pascal}
Mark Everingham, Luc Van~Gool, Christopher~KI Williams, John Winn, and Andrew
  Zisserman.
\newblock The pascal visual object classes (voc) challenge.
\newblock \emph{IJCV}, 2010.

\bibitem[Fan et~al.(2020)Fan, Zhang, Song, and Tan]{Fan_2020_CVPR}
Junsong Fan, Zhaoxiang Zhang, Chunfeng Song, and Tieniu Tan.
\newblock Learning integral objects with intra-class discriminator for
  weakly-supervised semantic segmentation.
\newblock In \emph{Proceedings of the IEEE/CVF Conference on Computer Vision
  and Pattern Recognition (CVPR)}, June 2020.

\bibitem[Fergus et~al.(2009)Fergus, Weiss, and Torralba]{fergus2009semi}
Rob Fergus, Yair Weiss, and Antonio Torralba.
\newblock Semi-supervised learning in gigantic image collections.
\newblock In \emph{Advances in neural information processing systems}, pp.\
  522--530, 2009.

\bibitem[Goldberger et~al.(2005)Goldberger, Hinton, Roweis, and
  Salakhutdinov]{goldberger2005neighbourhood}
Jacob Goldberger, Geoffrey~E Hinton, Sam~T Roweis, and Ruslan~R Salakhutdinov.
\newblock Neighbourhood components analysis.
\newblock In \emph{NIPS}, 2005.

\bibitem[He et~al.(2016)He, Zhang, Ren, and Sun]{he2016deep}
Kaiming He, Xiangyu Zhang, Shaoqing Ren, and Jian Sun.
\newblock Deep residual learning for image recognition.
\newblock In \emph{Proceedings of the IEEE conference on computer vision and
  pattern recognition}, pp.\  770--778, 2016.

\bibitem[Huang et~al.(2018)Huang, Wang, Wang, Liu, and Wang]{huang2018weakly}
Zilong Huang, Xinggang Wang, Jiasi Wang, Wenyu Liu, and Jingdong Wang.
\newblock Weakly-supervised semantic segmentation network with deep seeded
  region growing.
\newblock In \emph{Proceedings of the IEEE Conference on Computer Vision and
  Pattern Recognition}, pp.\  7014--7023, 2018.

\bibitem[Hwang et~al.(2019)Hwang, Yu, Shi, Collins, Yang, Zhang, and
  Chen]{hwang2019segsort}
Jyh-Jing Hwang, Stella~X Yu, Jianbo Shi, Maxwell~D Collins, Tien-Ju Yang, Xiao
  Zhang, and Liang-Chieh Chen.
\newblock Segsort: Segmentation by discriminative sorting of segments.
\newblock In \emph{ICCV}, 2019.

\bibitem[Joachims(2003)]{joachims2003transductive}
Thorsten Joachims.
\newblock Transductive learning via spectral graph partitioning.
\newblock In \emph{Proceedings of the 20th International Conference on Machine
  Learning (ICML-03)}, pp.\  290--297, 2003.

\bibitem[Khoreva et~al.(2017)Khoreva, Benenson, Hosang, Hein, and
  Schiele]{khoreva2017simple}
Anna Khoreva, Rodrigo Benenson, Jan Hosang, Matthias Hein, and Bernt Schiele.
\newblock Simple does it: Weakly supervised instance and semantic segmentation.
\newblock In \emph{Proceedings of the IEEE conference on computer vision and
  pattern recognition}, pp.\  876--885, 2017.

\bibitem[Kingma et~al.(2014)Kingma, Mohamed, Rezende, and
  Welling]{kingma2014semi}
Durk~P Kingma, Shakir Mohamed, Danilo~Jimenez Rezende, and Max Welling.
\newblock Semi-supervised learning with deep generative models.
\newblock In \emph{Advances in neural information processing systems}, pp.\
  3581--3589, 2014.

\bibitem[Kolesnikov \& Lampert(2016)Kolesnikov and Lampert]{kolesnikov2016seed}
Alexander Kolesnikov and Christoph~H Lampert.
\newblock Seed, expand and constrain: Three principles for weakly-supervised
  image segmentation.
\newblock In \emph{European Conference on Computer Vision}, pp.\  695--711.
  Springer, 2016.

\bibitem[Kr{\"a}henb{\"u}hl \& Koltun(2011)Kr{\"a}henb{\"u}hl and
  Koltun]{krahenbuhl2011efficient}
Philipp Kr{\"a}henb{\"u}hl and Vladlen Koltun.
\newblock Efficient inference in fully connected crfs with gaussian edge
  potentials.
\newblock In \emph{Advances in neural information processing systems}, pp.\
  109--117, 2011.

\bibitem[Lee et~al.(2019)Lee, Kim, Lee, Lee, and Yoon]{lee2019ficklenet}
Jungbeom Lee, Eunji Kim, Sungmin Lee, Jangho Lee, and Sungroh Yoon.
\newblock Ficklenet: Weakly and semi-supervised semantic image segmentation
  using stochastic inference.
\newblock In \emph{Proceedings of the IEEE conference on computer vision and
  pattern recognition}, pp.\  5267--5276, 2019.

\bibitem[Li et~al.(2018)Li, Wu, Peng, Ernst, and Fu]{li2018tell}
Kunpeng Li, Ziyan Wu, Kuan-Chuan Peng, Jan Ernst, and Yun Fu.
\newblock Tell me where to look: Guided attention inference network.
\newblock In \emph{Proceedings of the IEEE Conference on Computer Vision and
  Pattern Recognition}, pp.\  9215--9223, 2018.

\bibitem[Lin et~al.(2016)Lin, Dai, Jia, He, and Sun]{lin2016scribblesup}
Di~Lin, Jifeng Dai, Jiaya Jia, Kaiming He, and Jian Sun.
\newblock Scribblesup: Scribble-supervised convolutional networks for semantic
  segmentation.
\newblock In \emph{Proceedings of the IEEE Conference on Computer Vision and
  Pattern Recognition}, pp.\  3159--3167, 2016.

\bibitem[Lin et~al.(2014)Lin, Maire, Belongie, Hays, Perona, Ramanan,
  Doll{\'a}r, and Zitnick]{lin2014microsoft}
Tsung-Yi Lin, Michael Maire, Serge Belongie, James Hays, Pietro Perona, Deva
  Ramanan, Piotr Doll{\'a}r, and C~Lawrence Zitnick.
\newblock Microsoft coco: Common objects in context.
\newblock In \emph{European conference on computer vision}, pp.\  740--755.
  Springer, 2014.

\bibitem[Liu et~al.(2019)Liu, Wu, Hu, and Lin]{liu2019deep}
Bin Liu, Zhirong Wu, Han Hu, and Stephen Lin.
\newblock Deep metric transfer for label propagation with limited annotated
  data.
\newblock In \emph{Proceedings of the IEEE International Conference on Computer
  Vision Workshops}, pp.\  0--0, 2019.

\bibitem[Liu et~al.(2011)Liu, Yuen, and Torralba]{liu2011nonparametric}
Ce~Liu, Jenny Yuen, and Antonio Torralba.
\newblock Nonparametric scene parsing via label transfer.
\newblock \emph{PAMI}, 2011.

\bibitem[Miyato et~al.(2018)Miyato, Maeda, Koyama, and
  Ishii]{miyato2018virtual}
Takeru Miyato, Shin-ichi Maeda, Masanori Koyama, and Shin Ishii.
\newblock Virtual adversarial training: a regularization method for supervised
  and semi-supervised learning.
\newblock \emph{IEEE transactions on pattern analysis and machine
  intelligence}, 41\penalty0 (8):\penalty0 1979--1993, 2018.

\bibitem[Nystrom et~al.(2015)Nystrom, Levine, Roskies, and
  Scott]{nystrom2015bridges}
Nicholas~A Nystrom, Michael~J Levine, Ralph~Z Roskies, and J~Ray Scott.
\newblock Bridges: a uniquely flexible hpc resource for new communities and
  data analytics.
\newblock In \emph{Proceedings of the 2015 XSEDE Conference: Scientific
  Advancements Enabled by Enhanced Cyberinfrastructure}, pp.\  1--8, 2015.

\bibitem[Papandreou et~al.(2015)Papandreou, Chen, Murphy, and
  Yuille]{papandreou2015weakly}
George Papandreou, Liang-Chieh Chen, Kevin~P Murphy, and Alan~L Yuille.
\newblock Weakly-and semi-supervised learning of a deep convolutional network
  for semantic image segmentation.
\newblock In \emph{Proceedings of the IEEE international conference on computer
  vision}, pp.\  1742--1750, 2015.

\bibitem[Pathak et~al.(2015)Pathak, Krahenbuhl, and
  Darrell]{pathak2015constrained}
Deepak Pathak, Philipp Krahenbuhl, and Trevor Darrell.
\newblock Constrained convolutional neural networks for weakly supervised
  segmentation.
\newblock In \emph{Proceedings of the IEEE international conference on computer
  vision}, pp.\  1796--1804, 2015.

\bibitem[Rasmus et~al.(2015)Rasmus, Berglund, Honkala, Valpola, and
  Raiko]{rasmus2015semi}
Antti Rasmus, Mathias Berglund, Mikko Honkala, Harri Valpola, and Tapani Raiko.
\newblock Semi-supervised learning with ladder networks.
\newblock In \emph{Advances in neural information processing systems}, pp.\
  3546--3554, 2015.

\bibitem[Russell et~al.(2009)Russell, Efros, Sivic, Freeman, and
  Zisserman]{russell2009segmenting}
Bryan Russell, Alyosha Efros, Josef Sivic, Bill Freeman, and Andrew Zisserman.
\newblock Segmenting scenes by matching image composites.
\newblock In \emph{NIPS}, 2009.

\bibitem[Shimoda \& Yanai(2019)Shimoda and Yanai]{shimoda2019self}
Wataru Shimoda and Keiji Yanai.
\newblock Self-supervised difference detection for weakly-supervised semantic
  segmentation.
\newblock In \emph{Proceedings of the IEEE International Conference on Computer
  Vision}, pp.\  5208--5217, 2019.

\bibitem[Song et~al.(2019)Song, Huang, Ouyang, and Wang]{song2019box}
Chunfeng Song, Yan Huang, Wanli Ouyang, and Liang Wang.
\newblock Box-driven class-wise region masking and filling rate guided loss for
  weakly supervised semantic segmentation.
\newblock In \emph{Proceedings of the IEEE Conference on Computer Vision and
  Pattern Recognition}, pp.\  3136--3145, 2019.

\bibitem[Sun et~al.(2020)Sun, Wang, Dai, and Van~Gool]{sun2020mining}
Guolei Sun, Wenguan Wang, Jifeng Dai, and Luc Van~Gool.
\newblock Mining cross-image semantics for weakly supervised semantic
  segmentation.
\newblock In \emph{ECCV}, 2020.

\bibitem[Tang et~al.(2018{\natexlab{a}})Tang, Djelouah, Perazzi, Boykov, and
  Schroers]{tang2018normalized}
Meng Tang, Abdelaziz Djelouah, Federico Perazzi, Yuri Boykov, and Christopher
  Schroers.
\newblock Normalized cut loss for weakly-supervised cnn segmentation.
\newblock In \emph{Proceedings of the IEEE Conference on Computer Vision and
  Pattern Recognition}, pp.\  1818--1827, 2018{\natexlab{a}}.

\bibitem[Tang et~al.(2018{\natexlab{b}})Tang, Perazzi, Djelouah, Ben~Ayed,
  Schroers, and Boykov]{tang2018regularized}
Meng Tang, Federico Perazzi, Abdelaziz Djelouah, Ismail Ben~Ayed, Christopher
  Schroers, and Yuri Boykov.
\newblock On regularized losses for weakly-supervised cnn segmentation.
\newblock In \emph{Proceedings of the European Conference on Computer Vision
  (ECCV)}, pp.\  507--522, 2018{\natexlab{b}}.

\bibitem[Tarvainen \& Valpola(2017)Tarvainen and Valpola]{tarvainen2017mean}
Antti Tarvainen and Harri Valpola.
\newblock Mean teachers are better role models: Weight-averaged consistency
  targets improve semi-supervised deep learning results.
\newblock In \emph{Advances in neural information processing systems}, pp.\
  1195--1204, 2017.

\bibitem[Tighe \& Lazebnik(2010)Tighe and Lazebnik]{tighe2010superparsing}
Joseph Tighe and Svetlana Lazebnik.
\newblock Superparsing: scalable nonparametric image parsing with superpixels.
\newblock In \emph{ECCV}, 2010.

\bibitem[Towns et~al.(2014)Towns, Cockerill, Dahan, Foster, Gaither, Grimshaw,
  Hazlewood, Lathrop, Lifka, Peterson, et~al.]{towns2014xsede}
John Towns, Timothy Cockerill, Maytal Dahan, Ian Foster, Kelly Gaither, Andrew
  Grimshaw, Victor Hazlewood, Scott Lathrop, Dave Lifka, Gregory~D Peterson,
  et~al.
\newblock Xsede: accelerating scientific discovery.
\newblock \emph{Computing in science \& engineering}, 16\penalty0 (5):\penalty0
  62--74, 2014.

\bibitem[Wang et~al.(2019)Wang, Qi, Tang, Zhang, Wei, Li, and
  Zhang]{wang2019boundary}
Bin Wang, Guojun Qi, Sheng Tang, Tianzhu Zhang, Yunchao Wei, Linghui Li, and
  Yongdong Zhang.
\newblock Boundary perception guidance: A scribble-supervised semantic
  segmentation approach.
\newblock \emph{International Joint Conference on Artificial Intelligence},
  2019.

\bibitem[Wang et~al.(2020)Wang, Zhang, Kan, Shan, and Chen]{wang2020self}
Yude Wang, Jie Zhang, Meina Kan, Shiguang Shan, and Xilin Chen.
\newblock Self-supervised equivariant attention mechanism for weakly supervised
  semantic segmentation.
\newblock \emph{Proceedings of the IEEE International Conference on Computer
  Vision}, 2020.

\bibitem[Weston et~al.(2012)Weston, Ratle, Mobahi, and
  Collobert]{weston2012deep}
Jason Weston, Fr{\'e}d{\'e}ric Ratle, Hossein Mobahi, and Ronan Collobert.
\newblock Deep learning via semi-supervised embedding.
\newblock In \emph{Neural networks: Tricks of the trade}, pp.\  639--655.
  Springer, 2012.

\bibitem[Wu et~al.(2018{\natexlab{a}})Wu, Efros, and Yu]{wu2018improving}
Zhirong Wu, Alexei~A Efros, and Stella~X Yu.
\newblock Improving generalization via scalable neighborhood component
  analysis.
\newblock In \emph{Proceedings of the European Conference on Computer Vision
  (ECCV)}, pp.\  685--701, 2018{\natexlab{a}}.

\bibitem[Wu et~al.(2018{\natexlab{b}})Wu, Xiong, Yu, and
  Lin]{wu2018unsupervised}
Zhirong Wu, Yuanjun Xiong, Stella~X Yu, and Dahua Lin.
\newblock Unsupervised feature learning via non-parametric instance
  discrimination.
\newblock In \emph{Proceedings of the IEEE Conference on Computer Vision and
  Pattern Recognition}, pp.\  3733--3742, 2018{\natexlab{b}}.

\bibitem[Xie \& Tu(2015)Xie and Tu]{xie2015holistically}
Saining Xie and Zhuowen Tu.
\newblock Holistically-nested edge detection.
\newblock In \emph{Proceedings of the IEEE international conference on computer
  vision}, pp.\  1395--1403, 2015.

\bibitem[Xu et~al.(2015)Xu, Schwing, and Urtasun]{xu2015learning}
Jia Xu, Alexander~G Schwing, and Raquel Urtasun.
\newblock Learning to segment under various forms of weak supervision.
\newblock In \emph{Proceedings of the IEEE conference on computer vision and
  pattern recognition}, pp.\  3781--3790, 2015.

\bibitem[Yao \& Gong(2020)Yao and Gong]{yao2020saliency}
Qi~Yao and Xiaojin Gong.
\newblock Saliency guided self-attention network for weakly and semi-supervised
  semantic segmentation.
\newblock \emph{IEEE Access}, 8:\penalty0 14413--14423, 2020.

\bibitem[Zhang et~al.(2019)Zhang, Xiao, Wei, Sun, and
  Huang]{zhang2019reliability}
Bingfeng Zhang, Jimin Xiao, Yunchao Wei, Mingjie Sun, and Kaizhu Huang.
\newblock Reliability does matter: An end-to-end weakly supervised semantic
  segmentation approach.
\newblock \emph{arXiv preprint arXiv:1911.08039}, 2019.

\bibitem[Zhao et~al.(2017)Zhao, Shi, Qi, Wang, and Jia]{zhao2017pyramid}
Hengshuang Zhao, Jianping Shi, Xiaojuan Qi, Xiaogang Wang, and Jiaya Jia.
\newblock Pyramid scene parsing network.
\newblock In \emph{Proceedings of the IEEE conference on computer vision and
  pattern recognition}, pp.\  2881--2890, 2017.

\bibitem[Zhou et~al.(2016)Zhou, Khosla, Lapedriza, Oliva, and
  Torralba]{zhou2016learning}
Bolei Zhou, Aditya Khosla, Agata Lapedriza, Aude Oliva, and Antonio Torralba.
\newblock Learning deep features for discriminative localization.
\newblock In \emph{Proceedings of the IEEE conference on computer vision and
  pattern recognition}, pp.\  2921--2929, 2016.

\bibitem[Zhou et~al.(2004)Zhou, Bousquet, Lal, Weston, and
  Sch{\"o}lkopf]{zhou2004learning}
Dengyong Zhou, Olivier Bousquet, Thomas~N Lal, Jason Weston, and Bernhard
  Sch{\"o}lkopf.
\newblock Learning with local and global consistency.
\newblock In \emph{Advances in neural information processing systems}, pp.\
  321--328, 2004.

\end{thebibliography}
